\documentclass[10pt,twocolumn,letterpaper]{article}

\usepackage{cvpr}              

\usepackage{multirow}

\definecolor{cvprblue}{rgb}{0.21,0.49,0.74}
\usepackage[pagebackref,breaklinks,colorlinks,allcolors=cvprblue]{hyperref}
\usepackage{graphicx}
\usepackage{cuted}
\usepackage{capt-of}
\usepackage{blindtext}
\usepackage{graphicx}
\usepackage{caption}
\usepackage{overpic}
\usepackage{appendix}

\usepackage{xcolor}

\usepackage{tikz}
\usepackage[accsupp]{axessibility}  

\def\paperID{32864}
\def\confName{CVPR}
\def\confYear{2026}
\newcommand{\modelName}{\textit{HiFi-BRep}}

\title{HiFi-BRep: High-Fidelity Latent Representation for Robust B-Rep Generation}

\author{
Junhao Hou$^{1,*}$ \quad
Chenqi Luo$^{2,*}$ \quad
Pufan Wang$^{2}$ \quad
Jiaying Lu$^{1}$ \quad
Yusheng Liu$^{1}$ \quad \\
Feiwei Qin$^{2,\dagger}$ \quad
Meie Fang$^{3,\dagger}$ \quad
Kun Zhou$^{1,\dagger}$ \\
$^1$State Key Lab of CAD\&CG, Zhejiang University \quad \\
$^2$Hangzhou Dianzi University \quad
$^3$Guangzhou University \\
\small{$^{*}$Equal contribution \quad $^{\dagger}$Corresponding author}
}

\begin{document}

\twocolumn[{
\renewcommand\twocolumn[1][]{#1}
\maketitle

\thispagestyle{empty}
\begin{center}
     \includegraphics[width= 0.9\linewidth]{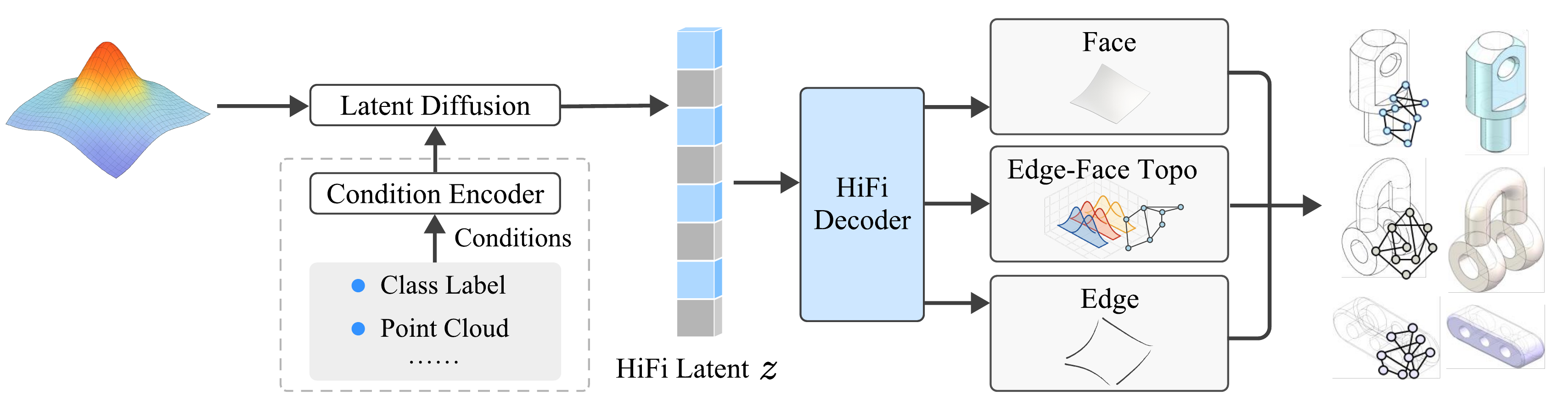}
     \captionof{figure}{
     \textbf{\modelName}~generates B-Rep models by jointly decoding geometry and topology from a high-fidelity latent space in a single stage, with key validity constraints embedded as differentiable objectives rather than deferred to post-processing. Benefiting from this unified representation, our approach supports both unconditional synthesis and generation conditioned on various inputs, such as class labels, point clouds, or images.}
    \label{fig:teaser}
\end{center}
}]

\begin{abstract}
Boundary representation (B-Rep) generation is a fundamental task in computer-aided design, yet the direct synthesis of high-fidelity and structurally valid B-Reps remains a major challenge. Existing deep generative methods suffer from two forms of brittleness: representation brittleness, caused by padding noise and feature contamination in the latent space, and generation brittleness, stemming from sequential error propagation and a train-inference mismatch due to non-differentiable validity enforcement. We propose \modelName, a novel framework that addresses these limitations through two synergistic contributions. First, a topology-aware encoder constructs a high-fidelity latent representation by eliminating padding via learnable queries and preventing feature contamination with topology-guided attention. Second, a single-stage decoder jointly predicts geometry and topology in parallel, embedding core manifold constraints as a differentiable learning objective. This design ensures mutual guidance between geometry and topology while avoiding cascaded errors. Extensive experiments show that \modelName~significantly outperforms state-of-the-art methods in both structural validity and geometric fidelity, providing a robust solution for high-quality B-Rep synthesis. Code and models are publicly available at \href{https://github.com/1nnoh/HiFi-BRep}{https://github.com/1nnoh/HiFi-BRep}. 

\end{abstract}

\section{Introduction}
\label{sec:intro}

Boundary representation (B-Rep) is the standard format in Computer-Aided Design (CAD), which encodes 3D shapes through parametric primitives (surfaces, curves, and vertices) and their topological connections. This representation enables precise modeling of complex geometries and serves as the foundation for industrial design, engineering, and manufacturing. Recent advances in deep generative models have demonstrated remarkable success in data-driven 3D content creation, such as meshes and point clouds. This naturally motivates the problem of B-Rep generation: can we extend this success to synthesize high-quality and structurally valid B-Reps?

Automatically generating complex and valid B-Reps is challenging, as it requires jointly modeling parametric geometry and topology under strict validity constraints. The hybrid continuous-discrete nature and stringent validity rules of B-Reps demand an exceptionally robust generative paradigm, since even minor errors can cascade and invalidate an entire model~\cite{liu2025hola}. However, existing approaches~\cite{jayaraman2023solidgen,xu2024brepgen,liu2025hola,li2025dtgbrepgen} exhibit fundamental brittleness, struggling to ensure both representational fidelity and structural validity simultaneously. This brittleness manifests in two primary forms: representation brittleness and generation brittleness.

First, \textbf{representation brittleness} stems from noise that corrupts the latent space. Early methods handle variable numbers of primitives through padding, which introduces statistical noise, destabilizes training, and scales poorly~\cite{jayaraman2023solidgen,xu2024brepgen}. While recent works incorporate topological priors~\cite{liu2025hola,li2025dtgbrepgen}, their encoding strategies can be suboptimal for generative tasks. For instance, propagating information across multi-hop topological neighbors~\cite{liu2025hola}, a design beneficial for analysis, risks contaminating features with irrelevant data when the goal is to predict direct, first-order adjacency, leading to mismatched inductive biases.

Second, \textbf{generation brittleness} arises from two design flaws in existing synthesis pipelines. The first is the reliance on sequential, cascaded generation processes, where geometry and topology are generated in multiple stages~\cite{jayaraman2023solidgen,xu2024brepgen,li2025dtgbrepgen,liu2025hola}. This one-way information flow prevents joint, bidirectional optimization of geometric and topological properties, locking the synthesis into suboptimal paths where early decisions become irreversible. Compounding this issue, many methods do not treat topological validity constraints as explicit learning objectives during training, instead deferring validity enforcement to non-differentiable post-processing. This creates a fundamental train-inference mismatch~\cite{xu2024brepgen,lee2025brepdiff}. Recent pioneering works offer partial solutions. For example, DTGBrepGen~\cite{li2025dtgbrepgen} successfully integrates topological validity constraints into its representation, but at the cost of a complex, multi-stage cascaded pipeline that is difficult to optimize. Conversely, HoLa~\cite{liu2025hola} simplifies the generation pipeline and ensures manifold edges through a clever local intersection paradigm. However, this design sacrifices expressive power, as it struggles to learn structures like multiple edges between two faces directly and hinders the learning of global topology. Thus, existing methods face inherent compromises, with each solution introducing distinct limitations. This motivates the need for a unified and parallel solution.

To address these challenges, we propose \textit{HiFi-BRep}, a novel framework for robust B-Rep generation. Our approach enhances robustness through two synergistic innovations: (1)~a topology-aware encoder that produces a high-fidelity latent representation without padding noise, and (2)~a validity-constrained single-stage decoder that embeds topological rules as differentiable inductive biases. The core idea is illustrated in Fig.~\ref{fig:overview}.

To mitigate representation brittleness, we design a high-fidelity B-Rep representation and a corresponding topology-aware encoder to eliminate representational noise and instability. Our design is guided by a key insight: an optimal representation should balance expressiveness with learnability while embedding validity constraints as learnable objectives. We therefore devise a compact yet sufficient representation composed of two geometric sequences (parametric B\'ezier curves and surfaces) and a single, explicit edge-face adjacency matrix, which serves as both a topology guide for encoding and a direct decoding target. This design offers several advantages. B\'ezier curves compactly represent both edges and their endpoint vertices, inherently satisfying the vertex connectivity constraint (each edge connects exactly two distinct vertices)~\cite{li2025dtgbrepgen}. Moreover, the global edge-face adjacency matrix naturally encodes manifold edges (each edge is shared by exactly two faces)~\cite{li2025dtgbrepgen} and accommodates complex structures, such as multiple edges between face pairs. Based on this representation, our encoder introduces two key mechanisms to ensure a high-fidelity latent space. First, to eliminate padding noise, it employs learnable queries to effectively aggregate variable-length sequences into a clean, fixed-length latent code. Second, to prevent feature contamination, it enforces topology-guided attention by strictly confining all cross-element interactions to adjacent pairs, using the explicit topology as a hard attention mask. This ensures that the encoder provides structurally informed features, enabling the decoder to reliably predict valid adjacency. Together, these mechanisms yield a latent representation that is high-fidelity, topology-aware, and validity-aware, providing a stable and expressive foundation for generation.

Building on this robust representation, we address generation brittleness with a novel validity-constrained single-stage decoder. This decoder tackles the challenge from two angles. First, it abandons the error-prone cascaded paradigm by predicting geometry and topology jointly and in parallel, enabling bidirectional optimization between them. Second, it enforces validity through a row-wise, two-peak learning objective that explicitly steers the model to satisfy the ``two faces per edge'' constraint during training. By design, this decoder synergizes with our latent representation: together, they provide manifold constraints, with vertex connectivity ensured by parametric curves and edge manifoldness enforced by the two-peak objective. This integrated design mitigates error propagation and aligns training with the physical constraints of manifold solids.

Our main contributions are summarized as follows:
\begin{itemize}
    \item We propose \modelName, a novel framework that tackles representation and generation brittleness in B-Rep synthesis.
    \item We introduce a compact B-Rep representation and topology-aware encoder that eliminates padding noise through learnable queries and prevents feature contamination via topology-guided attention, achieving high-fidelity latent representations.
    \item We design a single-stage validity-constrained decoder that jointly predicts geometry and topology in parallel, enforcing manifold constraints via a row-wise two-peak objective to promote valid B-Rep outputs.
    \item Extensive experiments show that \modelName~significantly outperforms state-of-the-art methods in both geometric fidelity and structural validity.
\end{itemize}


\section{Related Work}
\label{sec:related}
B-Rep generation requires addressing two fundamental challenges: learning robust representations that capture both geometry and topology, and designing generation methods that explicitly account for structural validity constraints.
This section reviews representation learning and generation methods for B-Reps. We also briefly discuss alternative CAD generation paradigms, including constructive solid geometry and procedural modeling.

\paragraph{B-Rep Representation Learning}
\label{sec:rw-representation}
Learning effective representations for B-Reps is crucial for downstream tasks such as shape classification, retrieval, and segmentation. Graph-based methods~\cite{lambourne2021brepnet,jayaraman2021uvnet,hou2023fus,cao2020graph,jones2021automate,willis2022joinable,jones2023self,bian2024hg} represent B-Reps as attributed graphs, where nodes encode geometric entities and edges encode topological relations, with some treating surfaces as nodes and others using heterogeneous graphs to encode different primitive types and their topological relationships. More recently, Transformer-based methods~\cite{lou2023brep,zou2025boundary,dai2025brepformer} have been proposed to leverage self-attention mechanisms for B-Rep learning, introducing specialized tokenization and embedding strategies to encode both geometric features and topological structures. However, these methods are primarily designed for analysis tasks, discarding reconstruction details and not explicitly modeling the validity of generated B-Reps.

\paragraph{B-Rep Generation Methods}
\label{sec:rw-generation}
B-Rep generation aims to jointly synthesize parametric geometry and discrete topology, yet the heterogeneity of primitives complicates unified representation and modeling. Early methods~\cite{jayaraman2023solidgen,xu2024brepgen} adopt multi-stage frameworks with hierarchical latent organizations, which lack compactness. SolidGen~\cite{jayaraman2023solidgen} employs cascaded autoregressive pipelines to predict geometry and topology sequentially, which propagates errors across stages and results in a fragile training paradigm. BRepGen~\cite{xu2024brepgen} relies on padding to handle variable-length primitives, introducing feature redundancy, and its reliance on post-processing for topology reconstruction precludes learning topological validity constraints. Recent works~\cite{liu2025hola,li2025dtgbrepgen,lee2025brepdiff,guo2025brepgiff,li2025stitch} seek to improve generation quality and robustness by incorporating validity priors or streamlining representations and pipelines. DTGBrepGen~\cite{li2025dtgbrepgen} integrates validity constraints but requires complex representations and multi-stage pipelines. While HoLa~\cite{liu2025hola} proposes a compact holistic representation and simplifies generation, it suffers from noise introduced by padding and limited global topology expressiveness. Despite adopting a single-stage generation process, BRepDiff~\cite{lee2025brepdiff} defers validity enforcement to post-processing, creating train-inference mismatch. 
Thus, while these approaches advance B-Rep generation, they struggle to simultaneously achieve compact and high-fidelity representation, joint geometry-topology generation, and differentiable enforcement of topological validity constraints.

\paragraph{Constructive Solid Geometry and Procedural Modeling}
\label{sec:rw-alternatives}
In addition to B-Reps, many CAD generators use procedural representations. Constructive solid geometry (CSG) combines parametric primitives via Boolean operations. Recent methods~\cite{sharma2018csgnet,kania2020ucsg,ren2021csg,yu2022capri,yu2023d} learn to reconstruct CSG trees with strong interpretability and editability, though expressiveness depends on the primitive library and redundancy arises from equivalent tree representations. Alternatively, construction history sequences (such as sketch-and-extrude operations) capture the step-by-step workflows recorded by CAD systems. Various approaches~\cite{wu2021deepcad,jones2020shapeassembly,xu2022skexgen,xu2023hierarchical,li2023secad,khan2024text2cad,chen2025img2cad,qin2025drawing2cad,li2025cad,wangtext} leveraging Transformers, large language models (LLMs), or agents have been proposed to generate such construction sequences, but are constrained by limited operation vocabularies. Moreover, datasets with complete construction histories are far smaller than large-scale B-Rep collections~\cite{willis2020fusion,koch2019abc}, further constraining pretraining scale and generalization.

\section{Method}
\label{sec:method}
In this section, we first provide an overview of our approach and then describe each component in detail.

\subsection{Overview}
Our goal is to learn a generative model for synthesizing diverse and valid B-Rep solids. We define a B-Rep solid $\mathcal{B}$ as a tuple of its geometric and topological components: a set of $n_f$ parametric faces $\mathbf{F}$, a set of $n_e$ parametric edges $\mathbf{E}$, and a binary edge-face adjacency matrix $\mathbf{A} \in \{0, 1\}^{n_e \times n_f}$. The core task is to model the complex joint probability distribution $p(\mathcal{B}) = p(\mathbf{F}, \mathbf{E}, \mathbf{A})$ while satisfying the strict manifold and watertight validity constraints inherent to solid models.

\begin{figure*}[t]
  \centering
  \begin{subfigure}[t]{0.9\textwidth}
    \centering
    \includegraphics[width=\textwidth]{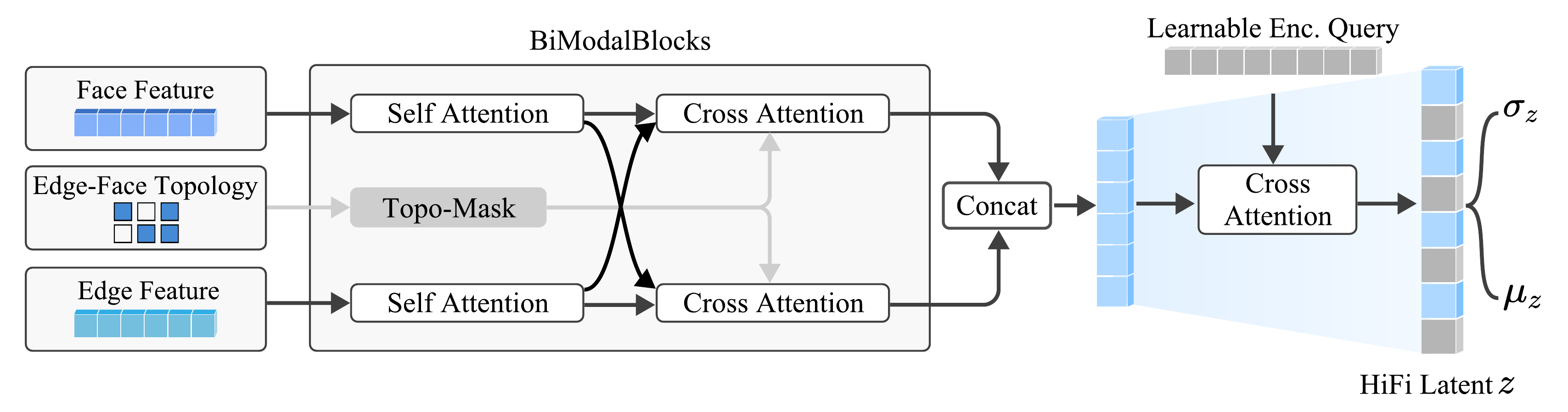} %
    \caption{Topology-aware dual-stream encoder.}
    \label{fig:enc}
  \end{subfigure}
  \vspace{6pt}

  \begin{subfigure}[t]{0.9\textwidth}
    \centering
    \includegraphics[width=\textwidth]{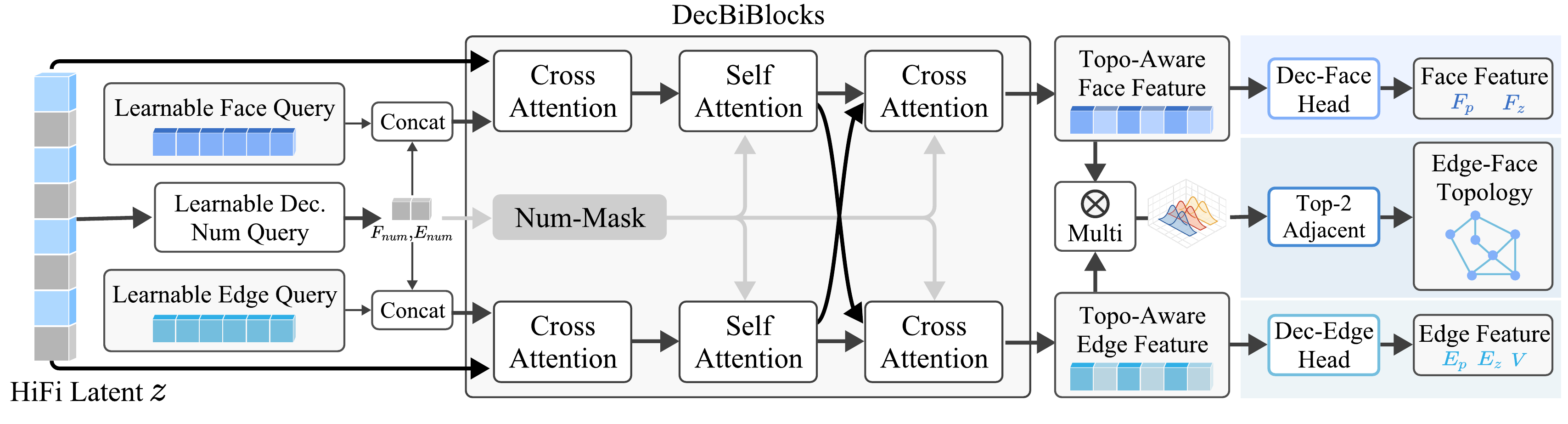} %
    \caption{Single-stage validity-constrained decoder.}
    \label{fig:dec}
  \end{subfigure}
  \caption{Overview of \modelName. (a)~Face and edge tokens undergo per-stream self-attention and cross-stream attention masked by edge-face incidence (Topo-Mask), then learnable queries pool them into a fixed-length latent. 
  (b)~The decoder first predicts face and edge counts to build hard padding masks, then updates learnable face and edge queries. Stacked DecBiBlocks then decode topology-aware geometry sequences. Geometry heads regress primitive parameters, while a topology solver head predicts adjacency with row-wise softmax and two-peak targets.}
  \label{fig:overview}
\end{figure*}

\modelName~addresses this problem with a two-stage pipeline built around a high-fidelity latent space.
As illustrated in Fig.~\ref{fig:overview}, we first train a variational autoencoder (VAE)~\cite{kingma2014vae} whose topology-aware encoder converts variable-length B-Rep inputs into a fixed-length latent sequence, and whose single-stage decoder reconstructs geometry and topology in parallel. 
The encoder targets \textit{representation brittleness} by restricting cross-stream interactions to topologically adjacent primitives and by pooling variable-length tokens with learnable queries instead of relying on padded global summaries.
The single-stage decoder targets \textit{generation brittleness} by jointly predicting face and edge geometric parameters and edge–face adjacency within one masked decoding pass, where a differentiable row-wise objective promotes the manifold prior that each edge belongs to exactly two faces.
Second, after learning a robust and structured latent space with the VAE, we train a latent diffusion model (LDM)~\cite{rombach2022ldm} on the resulting latent codes. This decoupled design separates representation learning from distribution modeling: the VAE captures the latent structure of complex B-Reps, while the LDM learns to generate samples in this well-behaved latent space. At inference, we draw a latent sample from the LDM and decode it with the pre-trained VAE decoder, enabling high-quality unconditional and conditional synthesis.

\begin{figure}[t]
\centering
\includegraphics[width=\columnwidth]{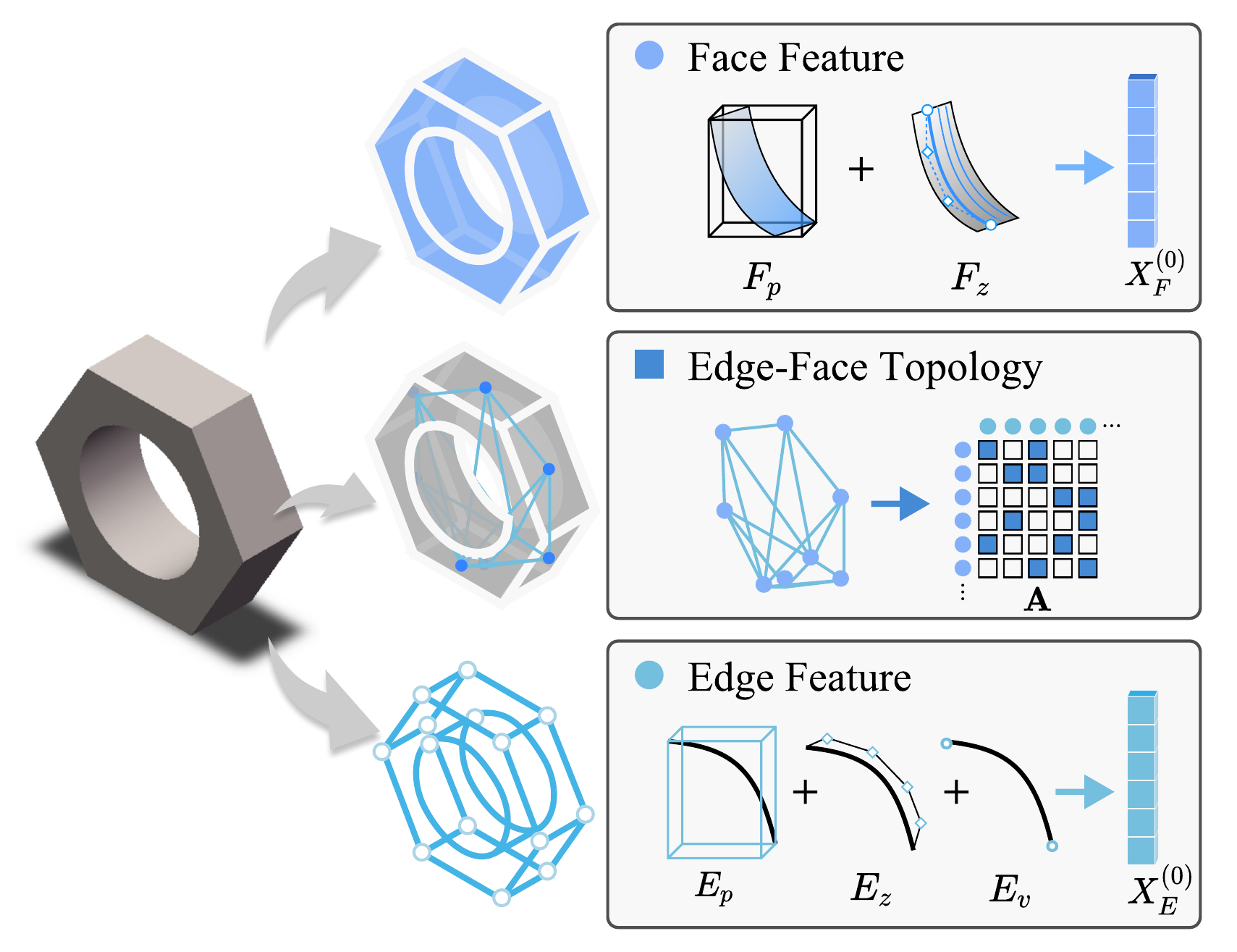}
\caption{Input B-Rep formulation. We construct unified primitive features by decomposing the shape into geometric and topological components. Specifically, we derive the face feature by summing the embeddings of the bounding box $F_p$ and B\'ezier control grid $F_z$. Similarly, the edge feature is formed by summing the embeddings of the bounding box $E_p$, control points $E_z$, and explicit endpoints $E_v$. The global structural connectivity is explicitly encoded via the edge-face incidence matrix.}
\label{fig:repr_overview}
\end{figure}
\subsection{Topology-Aware Dual-Stream Encoder}
\paragraph{Input B-Rep Formulation}
We adopt a compact parametric scheme that reduces representation noise while retaining smooth geometry and essential topological structure. As illustrated in Fig.~\ref{fig:repr_overview}, each face is modeled as a B\'ezier surface defined by a bounding box ($F_p$) and control points ($F_z$), while each edge is represented as a B\'ezier curve parameterized by a bounding box $E_p$, control points $E_z$, and explicit endpoints $E_v$ for precise global anchoring.
The edge-face topology is captured by an explicit incidence matrix ($\mathbf{A}$) for adjacency supervision.
For stability, we canonicalize each shape by lexicographically sorting primitives according to their bounding box centers. We then align the sequences to fixed budgets $(F_{\max}, E_{\max})$, defined as the maximum numbers of faces and edges in the dataset. In manifold solids, the number of edges $n_e$ scales linearly with the number of faces $n_f$. Therefore, these budgets provide sufficient capacity without unnecessary redundancy. Sequences shorter than these limits are padded and strictly masked in attention. 
Finally, we reconstruct the B-Rep from the predicted face and edge parameters together with the edge-face topology. Specifically, by merging spatially coincident endpoints into unique vertices, we establish the vertex-edge connectivity required to trace ordered per-face loops.


\paragraph{HiFi Latent Representation}
The encoder aims to produce a high-fidelity latent representation while mitigating two sources of representation brittleness: contamination from distant primitives and noise introduced by padding. To this end, we encode faces and edges in two separate streams and restrict cross-stream interaction to topologically adjacent pairs.
Let $X_E^{(0)}=\phi_E(E_p,E_z,\mathcal{V})\!\in\!\mathbb{R}^{n_e\times d}$ and $X_F^{(0)}=\phi_F(F_p,F_z)\!\in\!\mathbb{R}^{n_f\times d}$ denote the initial edge and face token embeddings, respectively, where $\phi_E(\cdot)$ and $\phi_F(\cdot)$ are shared token-wise MLPs (optionally augmented with learned positional encodings) that map the corresponding geometric parameters to $d$-dimensional features.
Here $n_e$ and $n_f$ are the numbers of valid edges and faces before padding, bounded by $E_{\max}$ and $F_{\max}$. Each BiModalBlock applies self-attention within the face and edge streams and bidirectional cross-attention between topologically adjacent face-edge pairs under a Topo-Mask:
\begin{align*}
\mathrm{Attn}(\mathbf{Q},\mathbf{K},\mathbf{V};\mathbf{S})
&= \mathrm{softmax}\!\Big(
\tfrac{\mathbf{Q}\mathbf{K}^\top}{\sqrt{d}} + \mathbf{S}
\Big)\mathbf{V},\\
\mathbf{S}[u,i] &=
\begin{cases}
0, & \mathbf{A}[u,i]=1,\\
-\infty, & \text{otherwise,}
\end{cases}
\end{align*}
where $u\!\in\![1,n_e]$ indexes edges and $i\!\in\![1,n_f]$ indexes faces. Stacking $L$ such blocks yields topology-aware edge and face features $X_E^{(L)}$ and $X_F^{(L)}$, which preserve local structural interactions while suppressing irrelevant cross-stream mixing.
We then pool the variable-length token set into a fixed-length latent sequence using $L_q$ learnable encoder queries $\mathcal{Q}_{\mathrm{enc}}\!\in\!\mathbb{R}^{L_q\times d}$. These queries attend to the concatenated features $[X_E^{(L)},X_F^{(L)}]$ with the corresponding key-padding masks, producing the HiFi latent $\mathcal{Z}\!\in\!\mathbb{R}^{L_q\times d}$. The VAE parameterizes this latent with $(\mu_{\mathcal{Z}},\log\sigma_{\mathcal{Z}}^2)$ for reparameterized sampling. This design enables topology-guided message passing during encoding and yields a clean fixed-length latent that serves as a stable interface for both single-stage decoding and latent diffusion.

\subsection{Single-Stage Validity-Constrained Decoder}
To address generation brittleness caused by cascaded error propagation and post-hoc repair, we decode primitive counts, geometry, and adjacency in a single stage, allowing topology and geometry to inform each other throughout generation (Fig.~\ref{fig:dec}). 
Starting from the HiFi latent $\mathcal{Z}$, two learnable count queries predict logits over $\{0,\dots,F_{\max}\}$ and $\{0,\dots,E_{\max}\}$ for the numbers of faces and edges. During training, the ground-truth counts $(n_f,n_e)$ define hard padding masks for all subsequent attention layers and prediction heads; at inference, these masks are determined by the predicted counts $(\hat n_f,\hat n_e)$. This count-first design resolves sequence-length ambiguity before decoding the geometric and topological primitives.
We then initialize face queries $\mathcal{Q}_F$ and edge queries $\mathcal{Q}_E$, and use a stack of DecBiBlocks to decode topology-aware face and edge features from the latent $\mathcal{Z}$. Each block consists of (i) cross-attention from the face and edge queries $(\mathcal{Q}_F,\mathcal{Q}_E)$ to $\mathcal{Z}$, (ii) masked self-attention within each stream, (iii) bidirectional cross-attention between the face and edge streams, and (iv) feed-forward networks, producing topology-aware decoder features $H_F$ and $H_E$. Geometry heads regress the face and edge control parameters from these features. For bounding boxes, we predict centers and sizes and convert them to corner coordinates through a differentiable mapping, using \texttt{softplus} to ensure positive box sizes. Edge endpoints are regressed directly.
To predict topology, we project $H_E$ and $H_F$ into a shared adjacency space, $U=\psi_e(H_E)\!\in\!\mathbb{R}^{E_{\max}\times d_{\text{adj}}}$ and $W=\psi_f(H_F)\!\in\!\mathbb{R}^{F_{\max}\times d_{\text{adj}}}$, and compute scaled bilinear scores $S=(UW^\top)/\sqrt{d_{\text{adj}}}$. 
A row-wise softmax over $S$ is supervised by a two-peak target distribution, which assigns equal probability mass to the two incident faces of each valid edge. At inference, for each valid edge, we select the two highest-scoring valid faces according to the predicted compatibility scores. Together, count prediction stabilizes the variable-length decoding process, while the compatibility head turns the ``two faces per edge'' manifold prior into a differentiable training objective. As a result, the decoder promotes structurally consistent geometry-topology predictions within a single decoding stage, rather than relying on cascaded generation or post-hoc repair.

\subsection{Training Objectives}
\label{sec:training_objectives}
We train the decoder with geometric reconstruction and row-wise adjacency objectives to encourage valid topology and watertightness.
Let the decoder outputs be the predicted counts $(\hat n_f,\hat n_e)$, geometric parameters $(\widehat{F}_z,\widehat{F}_p,\widehat{E}_z,\widehat{E}_p,\widehat{\mathcal{V}})$, and adjacency scores $S$. The masked geometric reconstruction loss is
\begin{align}
\mathcal{L}_{\text{geom}}
=\, &\mathrm{MSE}(\widehat{F}_z,F_z)
 + \mathrm{MSE}(\widehat{E}_z,E_z)
 + \mathrm{MSE}(\widehat{F}_p,F_p) \nonumber\\
 &+ \mathrm{MSE}(\widehat{E}_p,E_p)
 + \mathrm{MSE}(\widehat{\mathcal{V}},\mathcal{V}),
\end{align}
and the full objective is
\begin{align}
\mathcal{L}
=\, &\lambda_{\mathrm{KL}}\mathcal{L}_{\mathrm{KL}}
 + \lambda_{\mathrm{len}}\!\big[\mathrm{CE}(\hat n_f,n_f)+\mathrm{CE}(\hat n_e,n_e)\big] \nonumber\\
 &+ \lambda_{\mathrm{geom}}\mathcal{L}_{\text{geom}}
 + \lambda_{\mathrm{adj}}\,\mathcal{L}_{\mathrm{row\text{-}wise}}(S),
\end{align}
where $\mathcal{L}_{\mathrm{row\text{-}wise}}$ is the row-wise adjacency loss and all terms are computed only on valid (non-padded) slots. We keep label smoothing and loss weights fixed across all experiments.

\subsection{Latent Diffusion in the Unified Latent Space}
After learning the HiFi latent space with the VAE, we train a denoising diffusion probabilistic model (DDPM) on the resulting fixed-length latent codes. Let $z_0$ denote the serialized latent $\mathcal{Z}$. The forward process is $q(z_t\!\mid\!z_0)=\mathcal{N}(\sqrt{\bar\alpha_t}\,z_0,(1-\bar\alpha_t)I)$ and a Diffusion Transformer (DiT) denoiser $\epsilon_\theta(z_t,t\,;\,c)$ minimizes $\mathbb{E}\|\epsilon-\epsilon_\theta(z_t,t\,;\,c)\|_2^2$. 
Conditioning $c$ is optional and, when present, is injected through adaptive LayerNorm (adaLN), where a linear head maps $c$ to per-block scale and shift parameters $(\gamma,\beta)$. In our experiments, image conditions are encoded with a pretrained DINOv2~\cite{oquab2023dinov2}, while point-cloud conditions are encoded with a PointNet{++}~\cite{qi2017pointnet++}. At inference, we sample $z_T\!\sim\!\mathcal{N}(0,I)$ and iteratively denoise it under the chosen condition $c$ (or a null embedding for unconditional generation) to obtain $z_0$, which is then decoded once by the pretrained VAE decoder.

\subsection{Complexity and Implementation Notes}
Per-stream self-attention scales as $\mathcal{O}(F_{\max}^2D)$ and $\mathcal{O}(E_{\max}^2D)$. Topo-Mask cross-attention scales with allowed pairs $\|\!\mathbf{A}\!\|_0$, i.e., $\mathcal{O}(\|\!A\!\|_0D)$. Latent pooling is $\mathcal{O}(L_q(F_{\max}{+}E_{\max})D)$.
Using exactly one token per face and one per edge avoids half-edge duplication and face–face candidate enumeration, so the token budget scales with $F{+}E$. Since $E$ is typically of the same order as $F$ for manifold solids, runtime grows approximately linearly with shape complexity.
Architectural widths, attention heads, diffusion schedules, and mixed-precision settings are reported in the appendix.

\section{Experiments}
\label{sec:exp}

\subsection{Datasets and Evaluation Protocol}
We evaluate \modelName\ on DeepCAD~\cite{wu2021deepcad} and ABC~\cite{koch2019abc}, the two standard public benchmarks for B-Rep generation. Following prior work~\cite{xu2024brepgen,li2025dtgbrepgen}, we deduplicate the training data and cap complexity by face and edge counts to match the token budgets defined in Sec.~\ref{sec:method}. The resulting training sets contain 83{,}611 shapes for DeepCAD and 186{,}148 shapes for ABC.

We report both distributional fidelity and CAD-level validity metrics. Specifically, Coverage (COV), MMD with Chamfer Distance (MMD-CD), and Jensen--Shannon Divergence (JSD) evaluate how well the generated set matches the reference distribution, while Novel, Unique, Compilability, and Valid characterize sample diversity and structural correctness. Compilability measures whether a generated model can be exported as a STEP file~\cite{iso201410303} by OpenCascade, whereas Valid further requires the exported solid to be watertight and manifold-consistent. Their difference therefore measures how often a file-constructible sample fails full kernel-level validity. Across all experiments, we use the same architecture and training budget: encoder/decoder width $d\!=\!768$ with 6 encoder and 6 decoder blocks and $L_q\!=\!48$ learnable encoder queries, together with an 18-layer DiT for latent diffusion. The VAE and DiT contain 304.7M and 193.4M parameters, respectively, and are trained on $2{\times}$RTX\,4090 GPUs for 3{,}000 and 1{,}000 epochs.



\subsection{Unconditional Generation Results}
\label{sec:exp:main}
Tab.~\ref{tab:main} summarizes unconditional generation on DeepCAD and ABC. On DeepCAD, \modelName\ attains the highest \textbf{Validity} (72.20\%) and the lowest MMD\mbox{-}CD (1.05), while keeping COV close to the programmatic DeepCAD baseline; this supports our claim that embedding manifold constraints at decoding time improves structural soundness without sacrificing distributional fidelity. On ABC, DTGBrepGen achieves the best distributional alignment (highest COV and lowest MMD\mbox{-}CD/JSD), whereas \modelName\ delivers the highest \textbf{Validity} (32.66\%) with competitive MMD\mbox{-}CD/JSD. Across both datasets, \modelName\ exhibits a markedly smaller Compilability–Validity gap than the strongest baseline: on DeepCAD the gap is $90.38\!\to\!72.20$ (18.18) for \modelName\ versus $92.48\!\to\!43.20$ (49.28) for DTGBrepGen, and on ABC it is $35.61\!\to\!32.66$ (2.95) for \modelName\ versus $50.55\!\to\!24.88$ (25.67) for DTGBrepGen, indicating that our single-stage, validity-aware decoder produces not only compilable but also more frequently manifold-consistent solids.

\begin{table*}[t]
\centering
\caption{Unconditional generation on DeepCAD and ABC.
Best is \textbf{bold}, second-best is \underline{underlined}.
\emph{MMD-CD and JSD are $\times100$ (DeepCAD convention)}.}
\label{tab:main}
\resizebox{\textwidth}{!}{
\begin{tabular}{llccccccc}
\toprule
Dataset & Method & COV~($\uparrow$) & MMD-CD~($\downarrow$) & JSD~($\downarrow$) & Novel~($\uparrow$) & Unique~($\uparrow$) & Compilability~(\%, $\uparrow$) & Valid~(\%, $\uparrow$)\\
\midrule
\multirow{5}{*}{DeepCAD}
& DeepCAD & \textbf{76.67} & \underline{1.09} & \textbf{0.77} & 93.80 & 89.79 & 88.46 & \underline{68.20} \\
& BRepGen  & 47.03 & 1.51 & 3.12 & \underline{99.72} & \underline{99.18} & 20.91 & 20.76 \\
& BrepDiff & 45.03 & 1.32 & 2.39 & 99.27 & 99.10 & 63.69 & 63.69 \\
& DTGBrepGen & \underline{73.50} & \underline{1.06} & \underline{0.98} & \underline{99.79} & \textbf{99.33} & \textbf{92.48} & 43.20 \\
& \modelName\ (ours) & 70.40 & \textbf{1.05} & 1.72 & \textbf{99.81} & 99.15 & \underline{90.38} & \textbf{72.20} \\
\midrule
\multirow{4}{*}{ABC}
& BRepGen   & 34.73 & 2.08 & 5.72 & 99.68 & 99.05 & 20.77 & 20.19 \\
& BrepDiff  & 41.59 & 1.72 & 2.39 & \textbf{99.79} & 99.00 & 22.05 & 20.05 \\
& DTGBrepGen& \textbf{70.63} & \textbf{1.30} & \textbf{1.55} & \underline{99.73} & \underline{99.12} & \textbf{50.55} & \underline{24.88} \\
& \modelName\ (ours) & \underline{57.93} & \underline{1.45} & \underline{1.81} & \underline{99.73} & \textbf{99.20} & \underline{35.61} & \textbf{32.66} \\
\bottomrule
\end{tabular}}
\end{table*}

\begin{figure}[h]
\centering
\includegraphics[width=\columnwidth]
{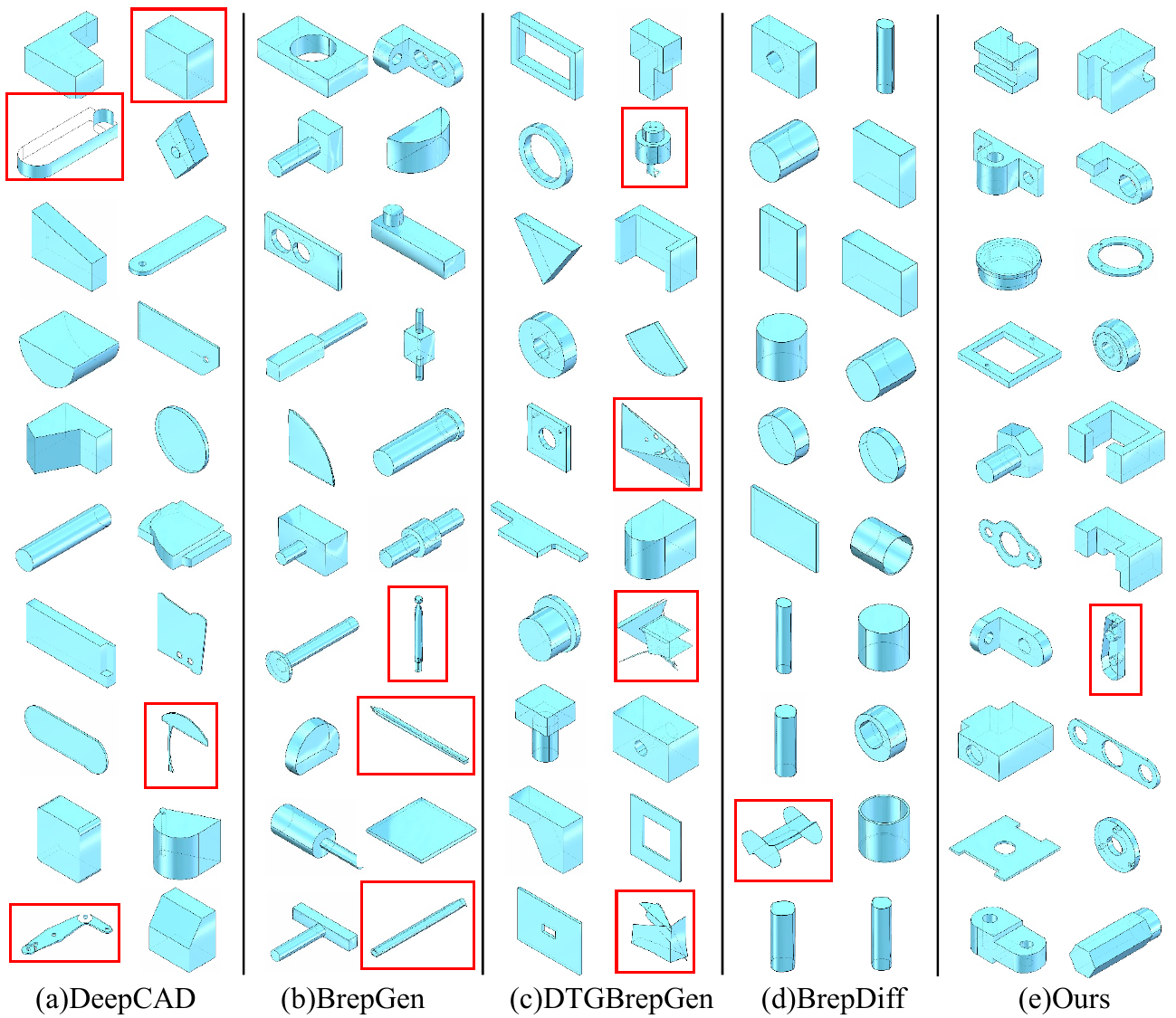}
\caption{Unconditional generations on DeepCAD.
Columns (a–e) show DeepCAD, BRepGen, DTGBrepGen, BrepDiff, and \modelName, respectively.
Red boxes highlight typical artifacts observed in baselines—open shells or missing faces around holes/fillets, non-manifold junctions, and degenerate thin plates/rods—while the rightmost column also marks a rare failure of our method for transparency.
Overall, \modelName\ yields more coherent solids with consistent edge–face incidence and fewer topology errors, consistent with its higher Validity.}
\label{fig:qual}
\end{figure}

We additionally note that qualitative comparisons in Fig.~\ref{fig:qual} mirror these trends: baselines often exhibit collapsed plates/rods, missing faces near holes/fillets, or non-manifold junctions, whereas \modelName\ typically preserves watertightness and edge–face consistency, aligning with its higher validity and smaller gap.

\begin{figure}[h]
\centering
\includegraphics[width=\columnwidth]{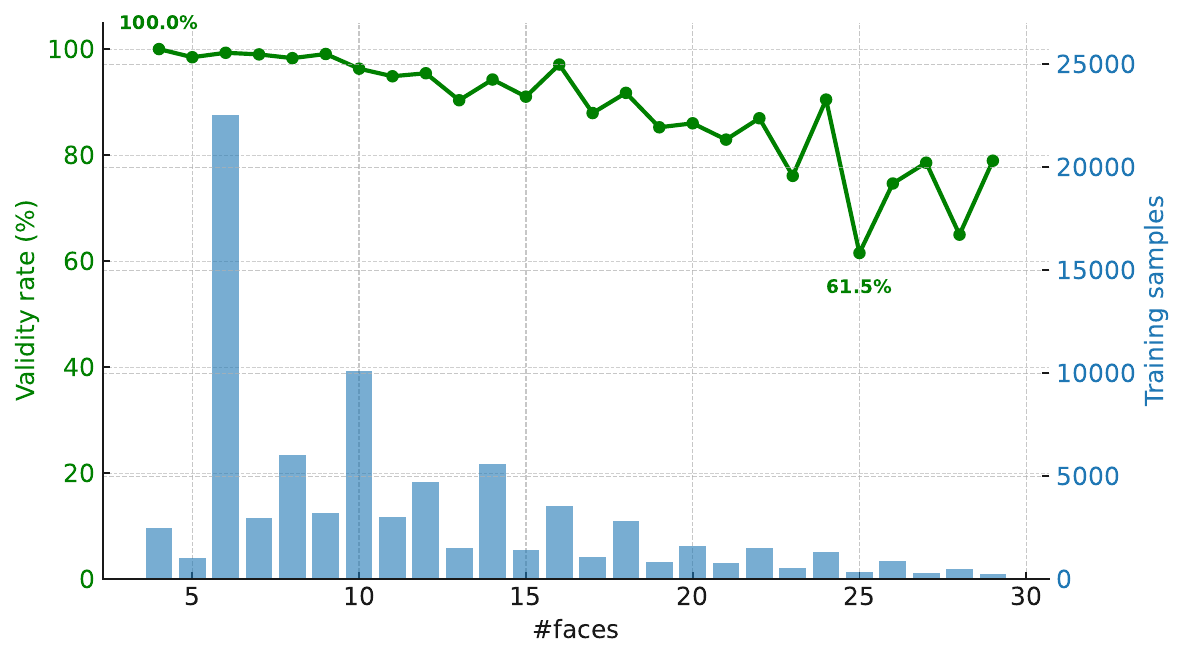}
\caption{Reconstruction validity by face count on DeepCAD.
Validity remains stable across common counts and $\geq61.5\%$ in high–face-count bins, evidencing robust generalization from the encoder.}
\label{fig:bins}
\end{figure}

\subsection{High-Fidelity Representation and Robust Reconstruction}
To assess whether the encoder yields a clean, topology-aware latent, we examine VAE reconstruction validity across exact face counts on DeepCAD (Fig.~\ref{fig:bins}).
Despite severe class imbalance (6–12 faces dominate, with a long tail up to 29), validity remains stable across prevalent counts and stays above $61.5\%$ in the rare high–face-count bins.
This pattern suggests that the padding-free latent and topology-constrained attention capture intrinsic topo–geometric couplings rather than overfitting frequent regimes.

\subsection{Ablation Studies}
\label{sec:exp:ablation}
We ablate four components that directly target representation or generation brittleness (Tab.~\ref{tab:ablation}). 
Here, \emph{Face Acc} denotes face-count accuracy, \emph{Edge Acc} denotes edge-count accuracy, and \emph{Adj Acc} denotes the accuracy of the edge–face incidence (adjacency) matrix. 
The largest drop (Valid $95.2\%\!\to\!69.3\%$, Adj Acc $97.5\%\!\to\!73.2\%$) occurs under the \emph{w/o two-stage} setting, where we first decode geometry using the VAE and then feed the decoded geometry into a separate topology predictor (architecturally similar to our encoder) to infer face–edge adjacency. 
This cascaded pipeline removes joint optimization, amplifies error accumulation from geometry into topology, and creates a misalignment between training and inference. 
Replacing the row-wise two-peak objective with independent BCE removes per-row competition; although inference still enforces top-2 selection, the scores become less well-calibrated and mis-rank candidates, increasing false positives within the top-2 and thereby reducing Adj Acc and Valid. 
Removing the encoder Topo-Mask dilutes cross-stream signal-to-noise, reducing Valid despite similar count accuracies. 
Canonicalization consistently reduces permutation-induced variance, yielding a small but stable gain.

\begin{table}[t]
\centering
\caption{Ablations on DeepCAD reconstruction. Metrics: \emph{Face Acc} = face-count accuracy; \emph{Edge Acc} = edge-count accuracy; \emph{Adj Acc} = edge–face incidence matrix accuracy. 
}
\label{tab:ablation}
\resizebox{\columnwidth}{!}{
\begin{tabular}{lcccc}
\toprule
Variant &  Face Acc.~(\%, $\uparrow$) & Edge Acc.~(\%, $\uparrow$) & Adj Acc.~(\%, $\uparrow$) & Valid~(\%, $\uparrow$) \\
\midrule
\modelName\ (full) &  \textbf{100.0} & \textbf{99.5} & \textbf{97.5} & \textbf{95.2} \\
\midrule
~\textit{w/o Topo-Mask (encoder)} & 99.3 & 98.7 & 92.7 & 89.5 \\
~\textit{w/o two-peak} & 99.3 & 98.5 & 90.4 & 87.2 \\
~\textit{w/o one-stage decoding} (geom$\rightarrow$adj) & 98.6 & 98.2 & 73.2 & 69.3 \\
\bottomrule
\end{tabular}}
\end{table}

\subsection{Efficiency and Runtime}
\label{sec:exp:eff}
We measure end-to-end latency on DeepCAD over \textbf{1,000} non-parallel generations. As shown in Tab.~\ref{tab:runtime_all_b}, \modelName attains the lowest total time at 3.83\,s/shape. Although our post-processing time (0.53\,s) is slightly longer than BRepGen's (0.32\,s), the single-stage pipeline avoids multi-pass decoding, yielding the fastest overall runtime. This modest overhead arises from our control-point parameterization, which requires a brief sampling and fitting step when exporting surfaces and curves. DTGBrepGen is particularly costly due to its post-processing that fits face and edge primitives (12.83\,s). BrepDiff is single-stage but relies on meshing and intersection to recover edges and vertices during post-processing (23.07\,s), which dominates its runtime. In summary, \modelName\ is \textbf{2.1$\times$} faster than BRepGen (8.09\,s), \textbf{6.2$\times$} faster than DTGBrepGen (23.55\,s), and \textbf{6.9$\times$} faster than BrepDiff (26.28\,s), while retaining the validity advantages reported in Sec.~\ref{sec:exp:main}.


\begin{table}[t]
\centering
\caption{Runtime comparison (seconds per shape) on the DeepCAD dataset. Results are averaged over 1000 runs}
\label{tab:runtime_all_b}
\resizebox{\linewidth}{!}{
\begin{tabular}{lcccc}
\toprule
Method & BRepGen & BrepDiff & DTGBrepGen & \textbf{\modelName} \\
\midrule
Total / Post (s) & 8.09 / 0.32 & 26.28 / 23.07 & 23.55 / 12.83 & \textbf{3.83 / 0.53} \\
\bottomrule
\end{tabular}}
\end{table}

Overall, \modelName\ consistently achieves higher structural validity, a smaller compilability–validity gap, robust reconstruction across topology scales, and faster inference, substantiating our design choice of embedding manifold constraints within a single-stage generative process.


\section{Limitations and Future Work}
\label{sec:limitations}
\begin{figure}[t]
\centering
\includegraphics[width=0.8\columnwidth]{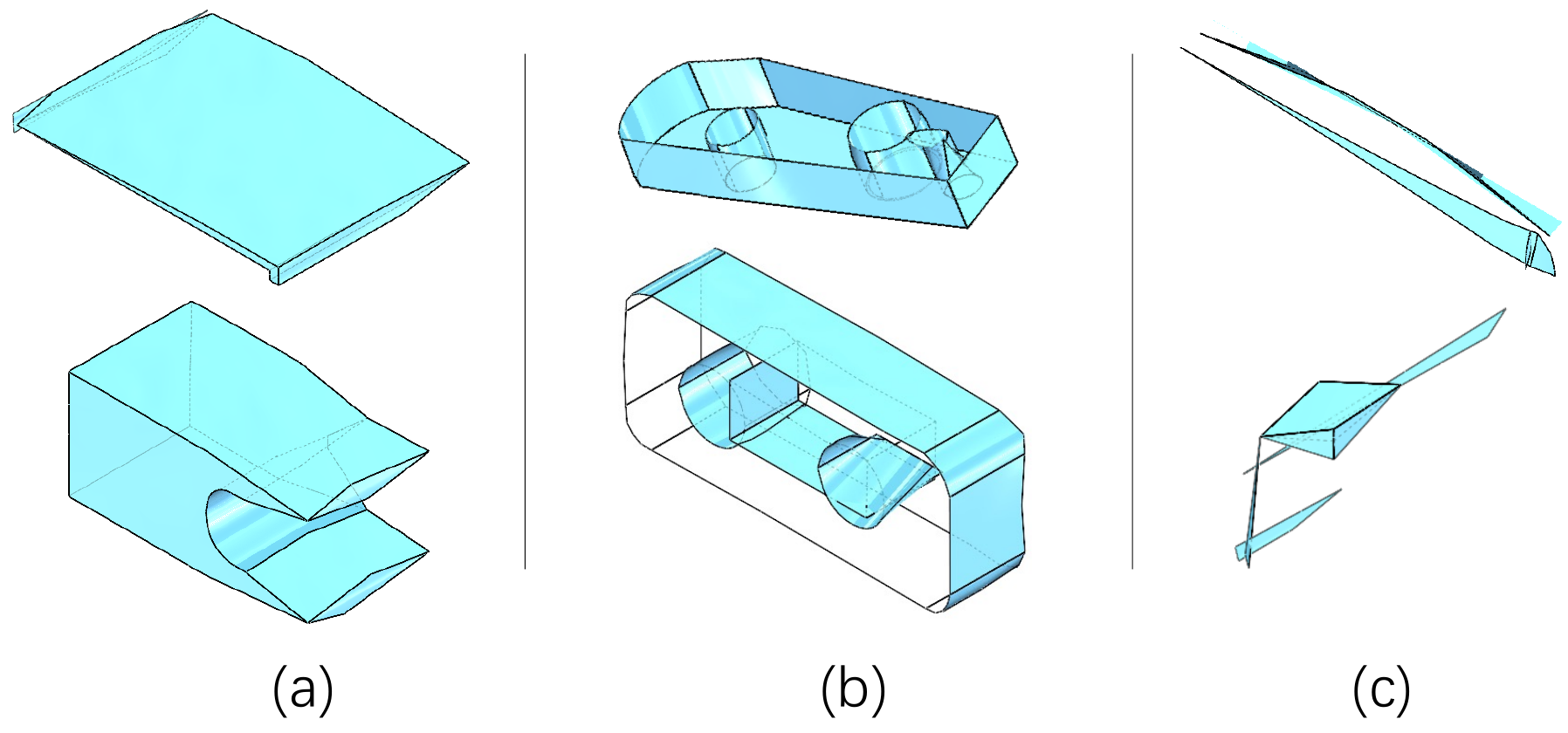}
\caption{Typical failure modes.
(a) \emph{Trimming disagreement / missing patch}: face count is correct, but decoded loops fail to form a valid trimmed region, so the kernel drops the face.
(b) \emph{Junction inconsistency / non-manifold edge}: T-junctions or duplicated segments after consolidation break manifold incidence.
(c) \emph{Degenerate geometry / sliver face}: ill-conditioned control points yield near-zero-area or self-intersecting patches.}
\label{fig:failure}
\end{figure}

This work targets closed, watertight B-Rep solids under a fixed face/edge budget and does not yet cover open-boundary parts, large assemblies, or non-manifold configurations. The one-shot decoder relies on accurate masks and tolerance choices during consolidation, and residual failures still occur as illustrated in Fig.~\ref{fig:failure}: trimming disagreement can remove a face even when the face count is predicted correctly, junction inconsistencies can appear after vertex consolidation despite top-2 adjacency selection, and ill-conditioned control points can yield sliver or self-intersecting patches. These cases explain part of the remaining Compilability–Validity gap and reflect that exact surface–curve intersection and trimming remain delegated to the CAD kernel.


Future work will focus on improving robustness and scalability without sacrificing the one-shot nature of the pipeline. Key directions include dynamic-capacity decoding with variable-length queries to handle long-tailed topology, and differentiable feasibility projections to suppress trimming and junction errors. We also plan to extend the framework to open-boundary models and assemblies, and enrich the representation with explicit vertex constraints and global consistency checks.

\section{Conclusion}
We presented \modelName, a robust framework for B-Rep generation that addresses the representation and generation brittleness in prior methods through two synergistic innovations. 
First, our topology-aware encoder with query-based pooling and topology-guided attention eliminates padding noise and feature contamination to achieve high-fidelity latent representations. 
Second, our single-stage decoder embeds validity constraints as differentiable objectives, enabling joint geometry-topology generation and mitigating cascaded errors. 
Extensive experiments demonstrate that \modelName~achieves state-of-the-art topology validity while maintaining strong consistency between compilability and manifold correctness, and delivers substantial efficiency gains over existing methods.  
Our work provides a robust and efficient foundation for neural B-Rep synthesis, supporting future research toward more reliable and scalable generative CAD systems.

\section*{Acknowledgments}
We thank the anonymous reviewers for their constructive comments.
This work is partially supported by the National Natural Science Foundation of China (No. 62421003), the Guangdong Basic and Applied Basic Research Foundation (No. 2026A1515011678), the Fundamental Research Funds for the Provincial Universities of Zhejiang (No. GK259909299001-006), and the XPLORER PRIZE.

{
    \small
    \bibliographystyle{ieeenat_fullname}
    \bibliography{main}

@String(CVPR= {IEEE Conf. Comput. Vis. Pattern Recog.})

@String(ICCV= {Int. Conf. Comput. Vis.})

@String(TOG= {ACM Trans. Graph.})

@String(ICLR = {Int. Conf. Learn. Represent.})

@String(CVPR  = {CVPR})

@String(ICCV  = {ICCV})

@String(TOG   = {ACM TOG})

@String(ICLR  = {ICLR})

@inproceedings{kingma2014vae,
  title     = {{Auto-Encoding Variational Bayes}},
  author    = {Kingma, Diederik P. and Welling, Max},
  booktitle = {Proc. Int. Conf. on Learning Representations (ICLR)},
  year      = {2014}
}

@inproceedings{rombach2022ldm,
  title     = {High-Resolution Image Synthesis with Latent Diffusion Models},
  author    = {Rombach, Robin and Blattmann, Andreas and Lorenz, Dominik and Esser, Patrick and Ommer, Björn},
  booktitle = {Proc. IEEE/CVF Conf. on Computer Vision and Pattern Recognition (CVPR)},
  pages={10684--10695},
  year={2022}
}

@inproceedings{peebles2023dit,
  title     = {Scalable Diffusion Models with {Transformers}},
  author    = {Peebles, William and Xie, Saining},
  booktitle = {Proc. IEEE/CVF Int. Conf. on Computer Vision (ICCV)},
  pages={4195--4205},
  year={2023}
}

@inproceedings{wu2021deepcad,
  title     = {{DeepCAD}: A Deep Generative Network for {Computer-Aided Design} Models},
  author    = {Wu, Rundi and Lambourne, Joseph and Willis, Karl and Jayaraman, Pradeep Kumar and Desai, Nishkrit and Sanghi, Aditya and Morris, Nigel},
  booktitle = {Proc. IEEE/CVF Int. Conf. on Computer Vision (ICCV)},
  pages={6772--6782},
  year      = {2021}
}

@article{jones2020shapeassembly,
  title     = {{ShapeAssembly}: Learning to Generate Programs for {3D} Shape Structure Synthesis},
  author    = {Jones, Ryan and Wu, Rundi and Willis, Karl and Desai, Nishkrit and Iorio, Federico and Chaudhuri, Siddhartha and Singh, Karan and Funkhouser, Thomas},
  journal   = {ACM Trans. Graph. (SIGGRAPH)},
  volume={39},
  number={6},
  pages={1--20},
  year={2020},
}

@article{khan2024text2cad,
  title={{Text2CAD}: Generating sequential {CAD} designs from beginner-to-expert level text prompts},
  author={Khan, Mohammad Sadil and Sinha, Sankalp and Uddin, Talha and Stricker, Didier and Ali, Sk Aziz and Afzal, Muhammad Zeshan},
  journal={Advances in Neural Information Processing Systems},
  volume={37},
  pages={7552--7579},
  year={2024}
}

@inproceedings{li2025cad,
  title={{CAD-Llama}: leveraging large language models for computer-aided design parametric {3D} model generation},
  author={Li, Jiahao and Ma, Weijian and Li, Xueyang and Lou, Yunzhong and Zhou, Guichun and Zhou, Xiangdong},
  booktitle={Proceedings of the Computer Vision and Pattern Recognition Conference},
  pages={18563--18573},
  year={2025}
}

@inproceedings{wangtext,
  title={{Text-to-CAD} Generation Through Infusing Visual Feedback in Large Language Models},
  author={Wang, Ruiyu and Yuan, Yu and Sun, Shizhao and Bian, Jiang},
  booktitle={Forty-second International Conference on Machine Learning},
  year={2025}
}

@inproceedings{lambourne2021brepnet,
  title={Brepnet: A topological message passing system for solid models},
  author={Lambourne, Joseph G and Willis, Karl DD and Jayaraman, Pradeep Kumar and Sanghi, Aditya and Meltzer, Peter and Shayani, Hooman},
  booktitle={Proceedings of the IEEE/CVF conference on computer vision and pattern recognition},
  pages={12773--12782},
  year={2021}
}

@inproceedings{jayaraman2021uvnet,
  title={{UV-Net}: Learning from Boundary Representations},
  author={Jayaraman, Pradeep Kumar and Sanghi, Aditya and Lambourne, Joseph G and Willis, Karl DD and Davies, Thomas and Shayani, Hooman and Morris, Nigel},
  booktitle={Proceedings of the IEEE/CVF conference on computer vision and pattern recognition},
  pages={11703--11712},
  year={2021}
}

@inproceedings{cao2020graph,
  title={Graph representation of {3D CAD} models for machining feature recognition with deep learning},
  author={Cao, Weijuan and Robinson, Trevor and Hua, Yang and Boussuge, Flavien and Colligan, Andrew R and Pan, Wanbin},
  booktitle={International design engineering technical conferences and computers and information in engineering conference},
  volume={84003},
  pages={V11AT11A003},
  year={2020},
  organization={American Society of Mechanical Engineers}
}

@article{jayaraman2023solidgen,
  title={{SolidGen}: An Autoregressive Model for Direct {B-Rep} Synthesis},
  author={Jayaraman, Pradeep Kumar and Lambourne, Joseph George and Desai, Nishkrit and Willis, Karl and Sanghi, Aditya and Morris, Nigel JW},
  journal={Transactions on Machine Learning Research},
  year={2023}
}

@article{xu2024brepgen,
  title={BrepGen: A B{-}rep Generative Diffusion Model with Structured Latent Geometry},
  author={Xu, Xiang and Lambourne, Joseph and Jayaraman, Pradeep and Wang, Zhengqing and Willis, Karl and Furukawa, Yasutaka},
  journal={ACM Transactions on Graphics (TOG)},
  volume={43},
  number={4},
  pages={1--14},
  year={2024},
  publisher={ACM New York, NY, USA}
}

@inproceedings{lee2025brepdiff,
  title={{BrepDiff}: Single-Stage {B-rep} Diffusion Model},
  author={Lee, Mingi and Zhang, Dongsu and Jambon, Cl{\'e}ment and Kim, Young Min},
  booktitle={Proceedings of the Special Interest Group on Computer Graphics and Interactive Techniques Conference Conference Papers},
  year={2025}
}

@inproceedings{guo2025brepgiff,
  title={BrepGiff: Lightweight Generation of Complex B-rep with 3D GAT Diffusion},
  author={Guo, Hao and Huang, Xiaoshui and Bai, Yunpeng and Gan, Hongping and Shi, Yilei and others},
  booktitle={Proceedings of the Computer Vision and Pattern Recognition Conference},
  pages={26587--26596},
  year={2025}
}

@inproceedings{li2025dtgbrepgen,
  title={{DTGBrepGen}: A Novel {B-rep} Generative Model through Decoupling Topology and Geometry},
  author={Li, Jing and Fu, Yihang and Chen, Falai},
  booktitle={Proceedings of the Computer Vision and Pattern Recognition Conference},
  pages={21438--21447},
  year={2025}
}

@article{liu2025hola,
  title={{HoLa}: B-rep generation using a holistic latent representation},
  author={Liu, Yilin and Xu, Duoteng and Yu, Xingyao and Xu, Xiang and Cohen-Or, Daniel and Zhang, Hao and Huang, Hui},
  journal={ACM Transactions on Graphics (TOG)},
  volume={44},
  number={4},
  pages={1--25},
  year={2025},
  publisher={ACM New York, NY, USA}
}

@inproceedings{li2025stitch,
  title={{Stitch-A-Shape}: Bottom-up Learning for {B-Rep} Generation},
  author={Li, Pu and Zhang, Wenhao and Chen, Jinglu and Yan, Dongming},
  booktitle={Proceedings of the Special Interest Group on Computer Graphics and Interactive Techniques Conference Conference Papers},
  pages={1--12},
  year={2025}
}

@article{hou2023fus,
  title={{FuS-GCN}: Efficient {B-rep} based graph convolutional networks for {3D-CAD} model classification and retrieval},
  author={Hou, Junhao and Luo, Chenqi and Qin, Feiwei and Shao, Yanli and Chen, Xiaxuan},
  journal={Advanced Engineering Informatics},
  volume={56},
  pages={102008},
  year={2023},
  publisher={Elsevier}
}

@article{willis2020fusion,
    title={Fusion 360 Gallery: A Dataset and Environment for Programmatic {CAD} Construction from Human Design Sequences},
    author={Karl D. D. Willis and Yewen Pu and Jieliang Luo and Hang Chu and Tao Du and Joseph G. Lambourne and Armando Solar-Lezama and Wojciech Matusik},
    journal={ACM Transactions on Graphics (TOG)},
    volume={40},
    number={4},
    year={2021},
    publisher={ACM New York, NY, USA}
}

@inproceedings{koch2019abc,
  title={{ABC}: A big {CAD} model dataset for geometric deep learning},
  author={Koch, Sebastian and Matveev, Albert and Jiang, Zhongshi and Williams, Francis and Artemov, Alexey and Burnaev, Evgeny and Alexa, Marc and Zorin, Denis and Panozzo, Daniele},
  booktitle={Proceedings of the IEEE/CVF conference on computer vision and pattern recognition},
  pages={9601--9611},
  year={2019}
}

@article{jones2021automate,
  title={Automate: A dataset and learning approach for automatic mating of {CAD} assemblies},
  author={Jones, Benjamin and Hildreth, Dalton and Chen, Duowen and Baran, Ilya and Kim, Vladimir G and Schulz, Adriana},
  journal={ACM Transactions on Graphics (TOG)},
  volume={40},
  number={6},
  pages={1--18},
  year={2021},
  publisher={ACM New York, NY, USA}
}

@inproceedings{willis2022joinable,
  title={Joinable: Learning bottom-up assembly of parametric {CAD} joints},
  author={Willis, Karl DD and Jayaraman, Pradeep Kumar and Chu, Hang and Tian, Yunsheng and Li, Yifei and Grandi, Daniele and Sanghi, Aditya and Tran, Linh and Lambourne, Joseph G and Solar-Lezama, Armando and others},
  booktitle={Proceedings of the IEEE/CVF conference on computer vision and pattern recognition},
  pages={15849--15860},
  year={2022}
}

@article{bian2024hg,
  title={{HG-CAD}: hierarchical graph learning for material prediction and recommendation in computer-aided design},
  author={Bian, Shijie and Grandi, Daniele and Liu, Tianyang and Jayaraman, Pradeep Kumar and Willis, Karl and Sadler, Elliot and Borijin, Bodia and Lu, Thomas and Otis, Richard and Ho, Nhut and others},
  journal={Journal of Computing and Information Science in Engineering},
  volume={24},
  number={1},
  pages={011007},
  year={2024},
  publisher={American Society of Mechanical Engineers}
}

@article{zou2025boundary,
  title={Boundary representation learning via Transformer},
  author={Zou, Qiang and Zhu, Lizhen},
  journal={Computer-Aided Design},
  volume = {189},
  pages = {103940},
  year = {2025},
  publisher={Elsevier}
}

@inproceedings{dai2025brepformer,
  title={{BRepFormer}: Transformer-Based {B-rep} Geometric Feature Recognition},
  author={Dai, Yongkang and Huang, Xiaoshui and Bai, Yunpeng and Guo, Hao and Gan, Hongping and Yang, Ling and Shi, Yilei},
  booktitle={Proceedings of the 2025 International Conference on Multimedia Retrieval},
  pages={155--163},
  year={2025}
}

@inproceedings{lou2023brep,
  title={{BRep-BERT}: Pre-training boundary representation {BERT} with sub-graph node contrastive learning},
  author={Lou, Yunzhong and Li, Xueyang and Chen, Haotian and Zhou, Xiangdong},
  booktitle={Proceedings of the 32nd ACM International Conference on Information and Knowledge Management},
  pages={1657--1666},
  year={2023}
}

@inproceedings{jones2023self,
  title={Self-Supervised Representation Learning for {CAD}},
  author={Jones, Benjamin T and Hu, Michael and Kodnongbua, Milin and Kim, Vladimir G and Schulz, Adriana},
  booktitle={2023 IEEE/CVF Conference on Computer Vision and Pattern Recognition (CVPR)},
  pages={21327--21336},
  year={2023},
  organization={IEEE}
}

@inproceedings{sharma2018csgnet,
  title={{CSGNet}: Neural shape parser for constructive solid geometry},
  author={Sharma, Gopal and Goyal, Rishabh and Liu, Difan and Kalogerakis, Evangelos and Maji, Subhransu},
  booktitle={Proceedings of the IEEE conference on computer vision and pattern recognition},
  pages={5515--5523},
  year={2018}
}

@article{kania2020ucsg,
  title={UCSG-NET-Unsupervised Discovering of Constructive Solid Geometry Tree},
  author={Kania, Kacper and Zieba, Maciej and Kajdanowicz, Tomasz},
  journal={Advances in Neural Information Processing Systems},
  volume={33},
  pages={8776--8786},
  year={2020}
}

@inproceedings{ren2021csg,
  title={{CSG-Stump}: A learning friendly {CSG-like} representation for interpretable shape parsing},
  author={Ren, Daxuan and Zheng, Jianmin and Cai, Jianfei and Li, Jiatong and Jiang, Haiyong and Cai, Zhongang and Zhang, Junzhe and Pan, Liang and Zhang, Mingyuan and Zhao, Haiyu and others},
  booktitle={Proceedings of the IEEE/CVF international conference on computer vision},
  pages={12478--12487},
  year={2021}
}

@inproceedings{yu2022capri,
  title={Capri-net: Learning compact {CAD} shapes with adaptive primitive assembly},
  author={Yu, Fenggen and Chen, Zhiqin and Li, Manyi and Sanghi, Aditya and Shayani, Hooman and Mahdavi-Amiri, Ali and Zhang, Hao},
  booktitle={Proceedings of the IEEE/CVF conference on computer vision and pattern recognition},
  pages={11768--11778},
  year={2022}
}

@article{yu2023d,
  title={{D$^2$CSG}: Unsupervised learning of compact {CSG} trees with dual complements and dropouts},
  author={Yu, Fenggen and Chen, Qimin and Tanveer, Maham and Mahdavi Amiri, Ali and Zhang, Hao},
  journal={Advances in Neural Information Processing Systems},
  volume={36},
  pages={22807--22819},
  year={2023}
}

@inproceedings{xu2022skexgen,
  title={{SkexGen}: Autoregressive Generation of {CAD} Construction Sequences with Disentangled Codebooks},
  author={Xu, Xiang and Willis, Karl DD and Lambourne, Joseph G and Cheng, Chin-Yi and Jayaraman, Pradeep Kumar and Furukawa, Yasutaka},
  booktitle={International Conference on Machine Learning},
  pages={24698--24724},
  year={2022},
  organization={PMLR}
}

@inproceedings{xu2023hierarchical,
  title={Hierarchical neural coding for controllable {CAD} model generation},
  author={Xu, Xiang and Jayaraman, Pradeep Kumar and Lambourne, Joseph G and Willis, Karl DD and Furukawa, Yasutaka},
  booktitle={Proceedings of the 40th International Conference on Machine Learning},
  pages={38443--38461},
  year={2023}
}

@inproceedings{li2023secad,
  title={{SECAD-Net}: Self-Supervised {CAD} Reconstruction by Learning Sketch-Extrude Operations},
  author={Li, Pu and Guo, Jianwei and Zhang, Xiaopeng and Yan, Dong-Ming},
  booktitle={Proceedings of the IEEE/CVF Conference on Computer Vision and Pattern Recognition},
  pages={16816--16826},
  year={2023}
}

@article{chen2025img2cad,
  title={{Img2CAD}: Conditioned {3-D CAD} model generation from single image with structured visual geometry},
  author={Chen, Tianrun and Yu, Chunan and Hu, Yuanqi and Li, Jing and Xu, Tao and Cao, Runlong and Zhu, Lanyun and Zang, Ying and Zhang, Yong and Li, Zejian and others},
  journal={IEEE Transactions on Industrial Informatics},
  year={2025},
  publisher={IEEE}
}

@inproceedings{qin2025drawing2cad,
  title={{Drawing2CAD}: Sequence-to-Sequence Learning for {CAD} Generation from Vector Drawings},
  author={Qin, Feiwei and Lu, Shichao and Hou, Junhao and Wang, Changmiao and Fang, Meie and Liu, Ligang},
  booktitle={Proceedings of the 33rd ACM International Conference on Multimedia},
  pages={10573--10582},
  year={2025}
}

@article{iso201410303,
  title={10303 Industrial Automation Systems and Integration—Product Data Representation and Exchange},
  author={ISO, ISO},
  journal={International Organization for Standardization: Geneva, Switzerland},
  year={2014}
}

@article{oquab2023dinov2,
  title={Dinov2: Learning robust visual features without supervision},
  author={Oquab, Maxime and Darcet, Timoth{\'e}e and Moutakanni, Th{\'e}o and Vo, Huy and Szafraniec, Marc and Khalidov, Vasil and Fernandez, Pierre and Haziza, Daniel and Massa, Francisco and El-Nouby, Alaaeldin and others},
  journal={arXiv preprint arXiv:2304.07193},
  year={2023}
}

@article{qi2017pointnet++,
  title={{PointNet++}: Deep hierarchical feature learning on point sets in a metric space},
  author={Qi, Charles Ruizhongtai and Yi, Li and Su, Hao and Guibas, Leonidas J},
  journal={Advances in neural information processing systems},
  volume={30},
  year={2017}
}

@inproceedings{chen2003visual,
  title={On visual similarity based 3D model retrieval},
  author={Chen, Ding-Yun and Tian, Xiao-Pei and Shen, Yu-Te and Ouhyoung, Ming},
  booktitle={Computer graphics forum},
  volume={22},
  number={3},
  pages={223--232},
  year={2003},
  organization={Wiley Online Library}
}

@article{song2020denoising,
  title={Denoising diffusion implicit models},
  author={Song, Jiaming and Meng, Chenlin and Ermon, Stefano},
  journal={arXiv preprint arXiv:2010.02502},
  year={2020}
}
}

\clearpage
\appendix
\clearpage
\setcounter{page}{1}
\maketitlesupplementary

\section{Implementation Details}
This section details the training configurations for our model. Unless stated otherwise, the optimizer, scheduler, and precision settings apply to both the VAE and the DiT components. All attention modules use $12$ heads by default. We train in mixed precision (\texttt{bfloat16}) and perform inference in \texttt{float32}.
The latent diffusion model is trained with DDPM for $1{,}000$ diffusion steps using \texttt{beta\_start}$=1\mathrm{e}{-4}$, \texttt{beta\_end}$=2\mathrm{e}{-2}$, the \texttt{squaredcos\_cap\_v2} schedule, and \texttt{prediction\_type} set to \texttt{sample}. At sampling time, we adopt DDIM~\cite{song2020denoising} with $400$ steps. Optimization uses AdamW with learning rate $1\mathrm{e}{-4}$, weight decay $1\mathrm{e}{-2}$, $(\beta_1,\beta_2)=(0.9,0.99)$, and $\epsilon=1\mathrm{e}{-8}$. For the objective in Sec.~\ref{sec:training_objectives}, we set the loss weights to $(\lambda_{\mathrm{KL}},\lambda_{\mathrm{len}},\lambda_{\mathrm{geom}},\lambda_{\mathrm{adj}})=(5\times10^{-5},\,1,\,25,\,5)$. The learning rate warms up linearly over the first $10$\% of updates from $0.01\times$ the base rate to the base rate, and then follows a cosine decay back to $0.01\times$ the base rate for the remaining steps.

\section{Novelty Verification}
\begin{figure*}[t]
\centering
\includegraphics[width=\textwidth]{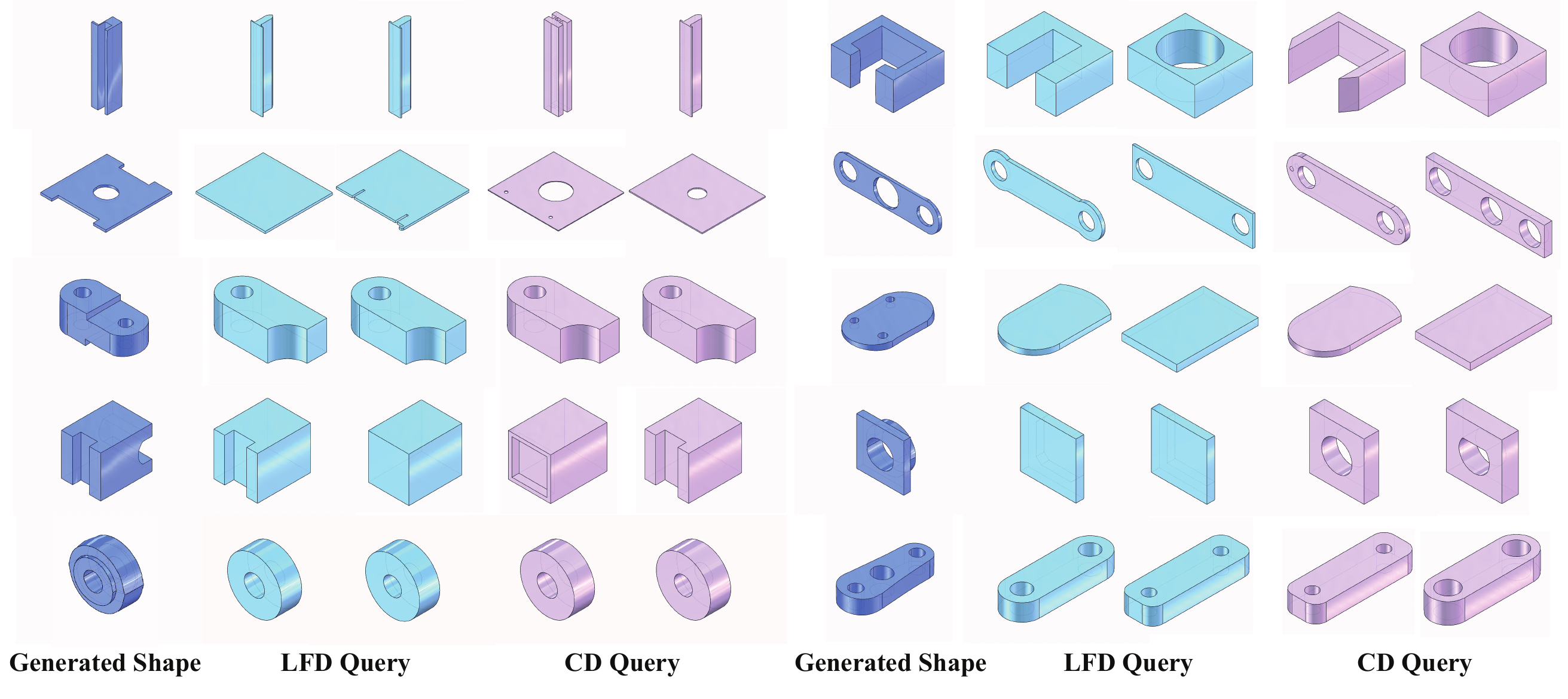}
\caption{\textbf{Retrieval{-}based novelty check.}
For each generated shape (left in each triplet), we show its two nearest neighbors from the training set retrieved by Chamfer Distance (CD) and Light Field Distance (LFD). Despite proximity under both metrics, the generated results display distinct geometric features, supporting that \modelName\ does not simply memorize training examples.}
\label{fig:retrieval_vis}
\end{figure*}

To assess whether \modelName\ synthesizes genuinely new shapes rather than recalling training instances, we perform a retrieval–based novelty check following the protocol of DTGBrepGen~\cite{li2025dtgbrepgen}. For each of 500 randomly generated B{-}reps, we compute Chamfer Distance (CD) on uniformly sampled surfaces and Light Field Distance (LFD)~\cite{chen2003visual} on multi{-}view renderings against the entire training corpus, and retrieve the two nearest neighbors under each metric. As illustrated in Fig.~\ref{fig:retrieval_vis}, the retrieved training shapes remain close under CD/LFD yet exhibit clear differences in patch layout, hole patterns, and junction geometry compared with our generations. These side{-}by{-}side comparisons indicate that the model is not merely reproducing training exemplars. Instead, the single{-}stage validity{-}aware decoding paired with the high{-}fidelity latent supports novel but plausible variations within the data manifold.

\section{Conditional Generation}
We study conditional generation to test whether our model can follow guidance while keeping valid topology. Conditions are injected in the diffusion denoiser with Adaptive LayerNorm (adaLN) as in DiT~\cite{peebles2023dit}. Each condition is first encoded to a vector and then used by adaLN to steer the denoising process.

We use the Furniture dataset~\cite{xu2024brepgen} for class conditioning because it has ten balanced categories with clear visual differences. For class labels, we map each of the ten categories to a 768-dimensional embedding vector. We use CADNet40~\cite{hou2023fus} for all other conditional generation experiments since it contains realistic industrial parts and supports practical control tasks with a moderate data size. For point clouds, we train a PointNet{++}~\cite{qi2017pointnet++} encoder to produce a 768-dimensional feature from 2{,}048 points. For partial point clouds, we drop approximately 30\% of points in a local region to mimic missing scans and use the same encoder. For single view images and wireframe sketches, we use a pretrained DINOv2~\cite{oquab2023dinov2} to obtain a 1{,}024-dimensional feature and project it to 768 dimensions. For multi view images, we encode each view with DINOv2, add a simple view position tag, average the view features, and project to 768 dimensions. We fine-tune the ABC-pretrained VAE on Furniture and CADNet40 for 300 epochs each, followed by training a dataset-specific DiT for 300 epochs. Other settings and splits follow the main training. The Furniture training set has 1{,}440 shapes and the CADNet40 training set has 7{,}394 shapes. All visual results use test set conditions.

\begin{figure*}[t]
\centering
\includegraphics[width=\textwidth]{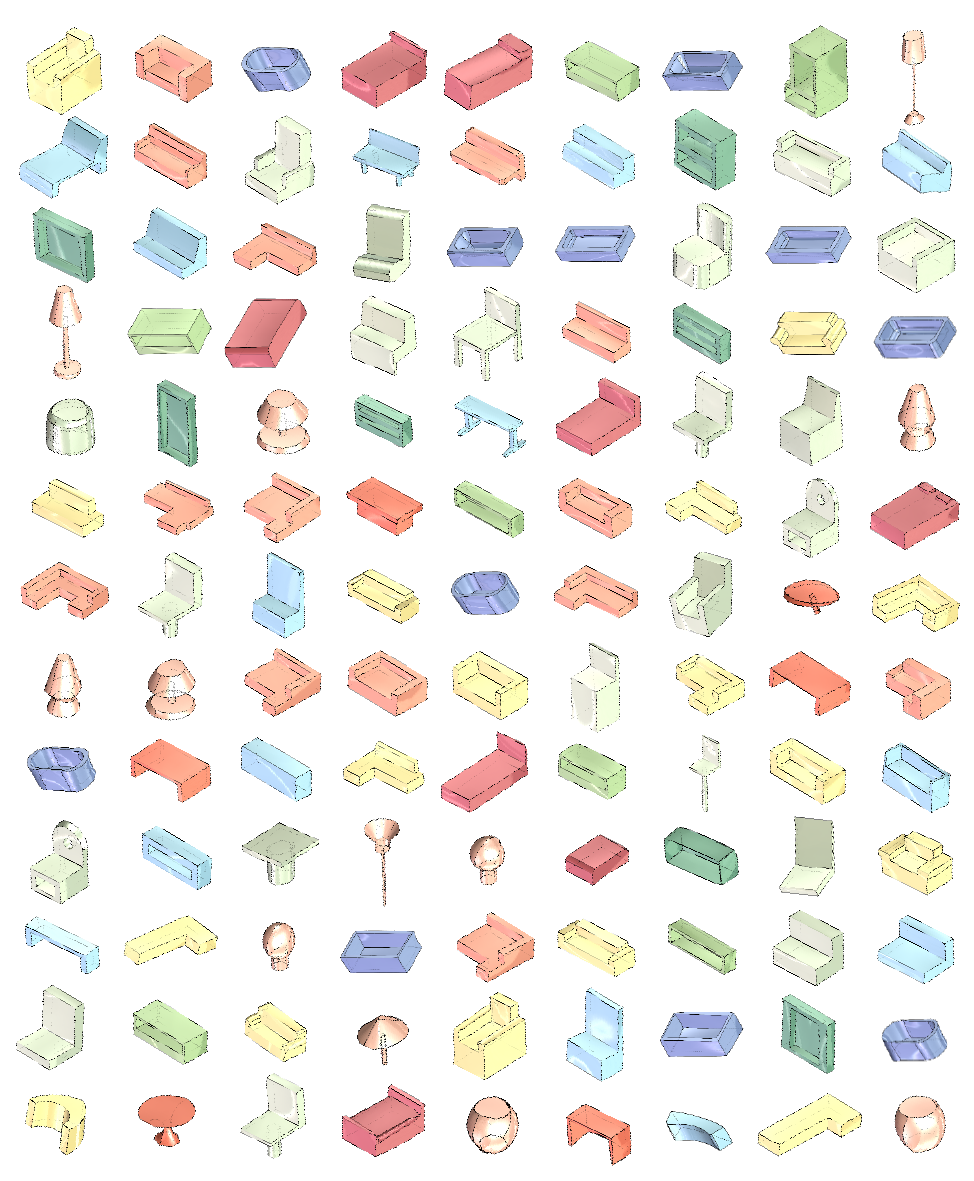}
\caption{\textbf{Class-conditioned generation on Furniture.}
Colors indicate categories (ten classes). Samples from different categories are interleaved, showing class-specific traits and intra-class diversity.}
\label{fig:cond_class}
\end{figure*}

\paragraph{Class-conditioned generation}
Fig.~\ref{fig:cond_class} shows the results on Furniture. The generations express category specific structures while keeping variations within each class.

\begin{figure*}[t]
\centering
\includegraphics[width=\textwidth]{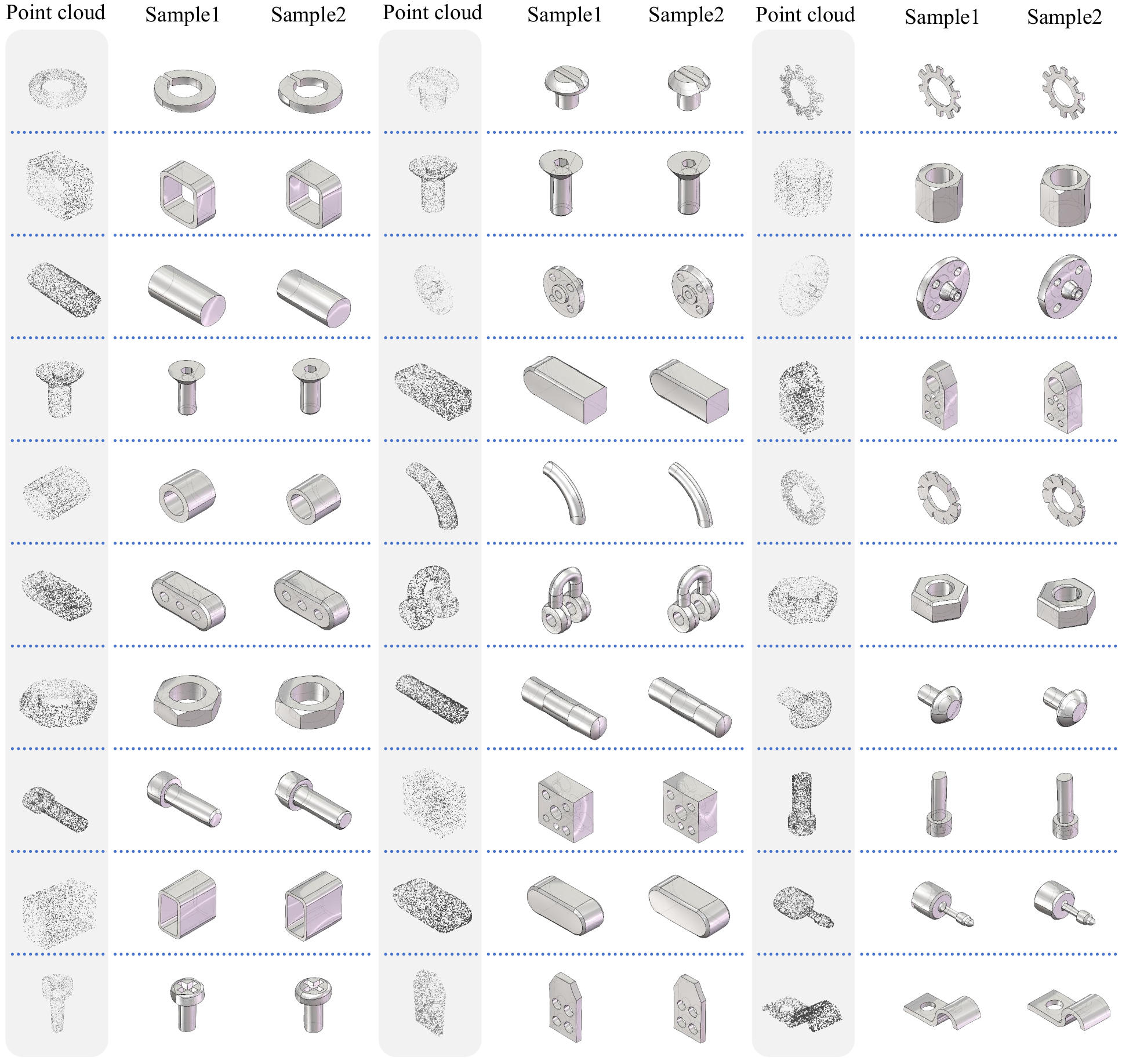}
\caption{\textbf{Point cloud-conditioned generation on CADNet40.}
Given a 2{,}048 point input, the model produces solids that match the global shape while allowing small variations.}
\label{fig:cond_pc}
\end{figure*}

\paragraph{Point cloud-conditioned generation}
Fig.~\ref{fig:cond_pc} shows the results on CADNet40. The outputs follow the geometry of the input points and preserve key features such as holes and fillets.

\begin{figure*}[t]
\centering
\includegraphics[width=\textwidth]{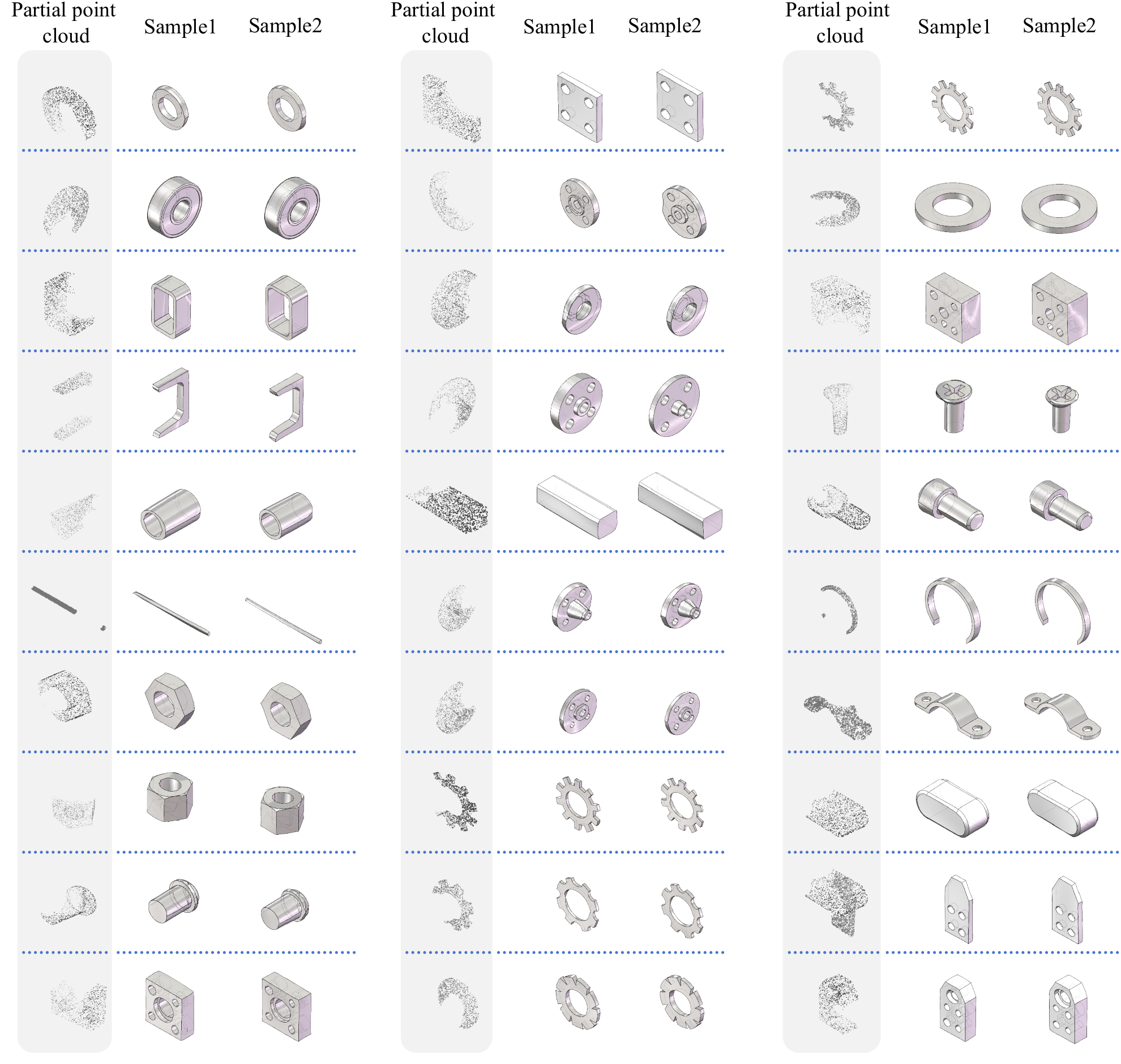}
\caption{\textbf{Partial point cloud-conditioned generation.}
The model generates plausible completions for missing regions while respecting visible geometry.}
\label{fig:cond_pc_partial}
\end{figure*}

\paragraph{Partial point cloud-conditioned generation}
Fig.~\ref{fig:cond_pc_partial} shows the results when the input point clouds have missing local regions. The generations complete the absent areas and keep the observed parts consistent with the input.

\begin{figure*}[t]
\centering
\includegraphics[width=\textwidth]{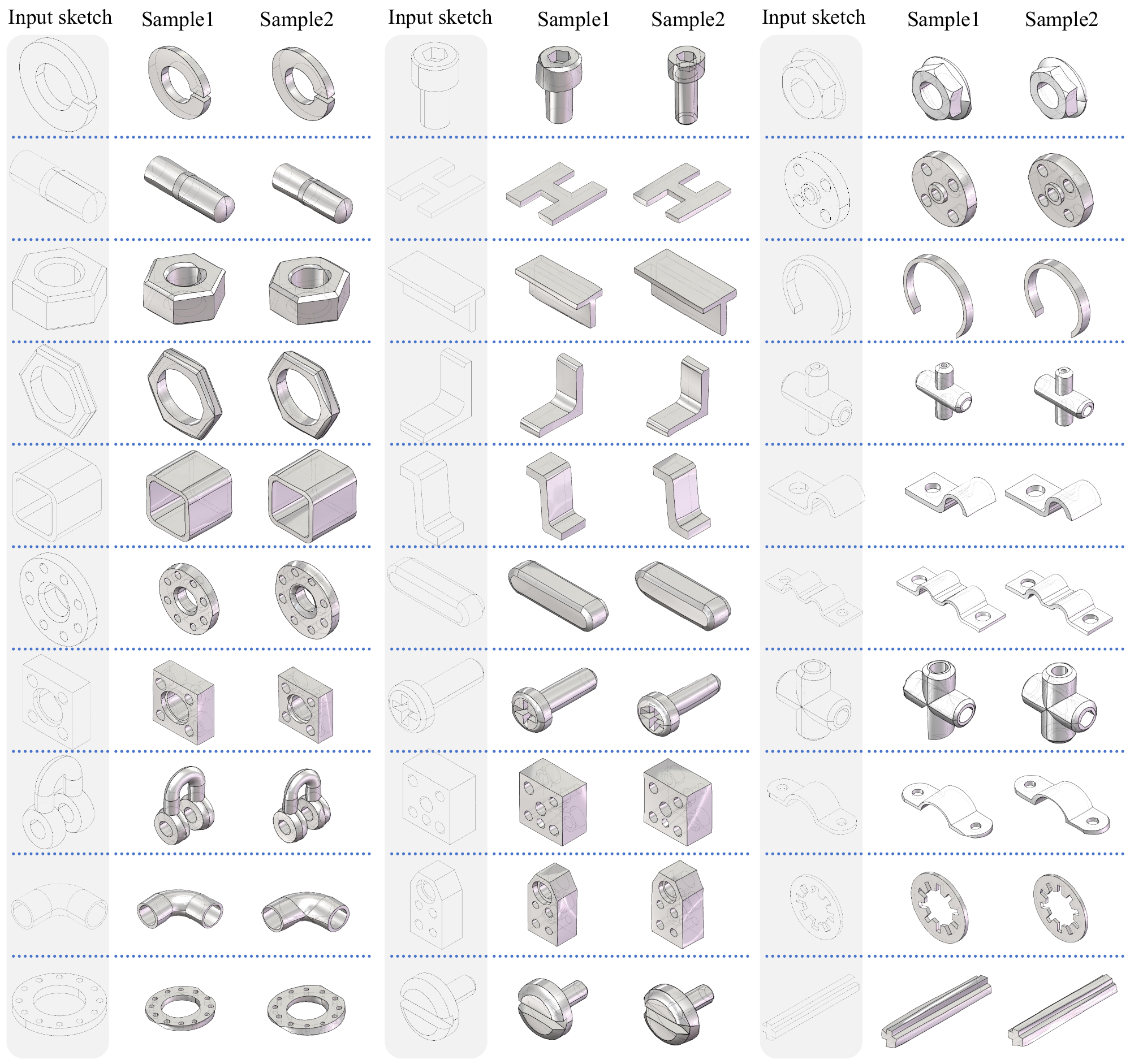}
\caption{\textbf{Sketch-conditioned generation.} 
Given a sketch, the model generates solids that respect the sketch while maintaining topological consistency.}
\label{fig:cond_wire}
\end{figure*}

\paragraph{Sketch-conditioned generation}
Fig.~\ref{fig:cond_wire} shows the results from wireframe sketches. The outputs align with the silhouette and recover coherent faces and edges.

\begin{figure*}[h]
\centering
\includegraphics[width=\textwidth]{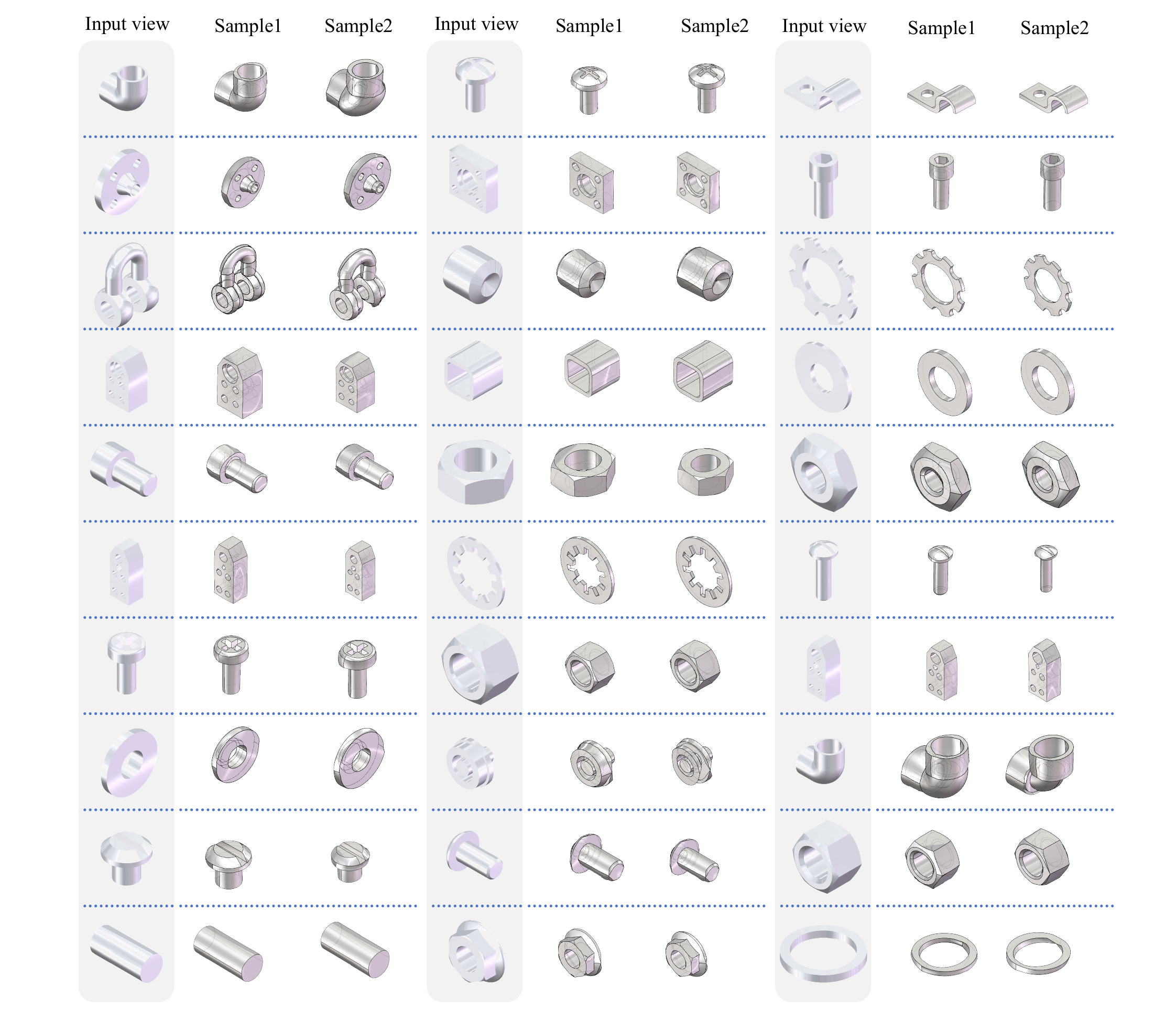}
\caption{\textbf{Single-view conditioned generation.}
For each input image, we present two generated candidates, at least one of which consistently aligns with the conditioning view while capturing the overall shape with minor variations.}
\label{fig:cond_single}
\end{figure*}

\paragraph{Single-view conditioned generation}
Fig.~\ref{fig:cond_single} illustrates single-view conditioning. For each input, we sample two candidates, at least one of which is view-consistent, while both typically capture the global shape without multi-view cues.

\begin{figure*}[t]
\centering
\includegraphics[width=\textwidth]{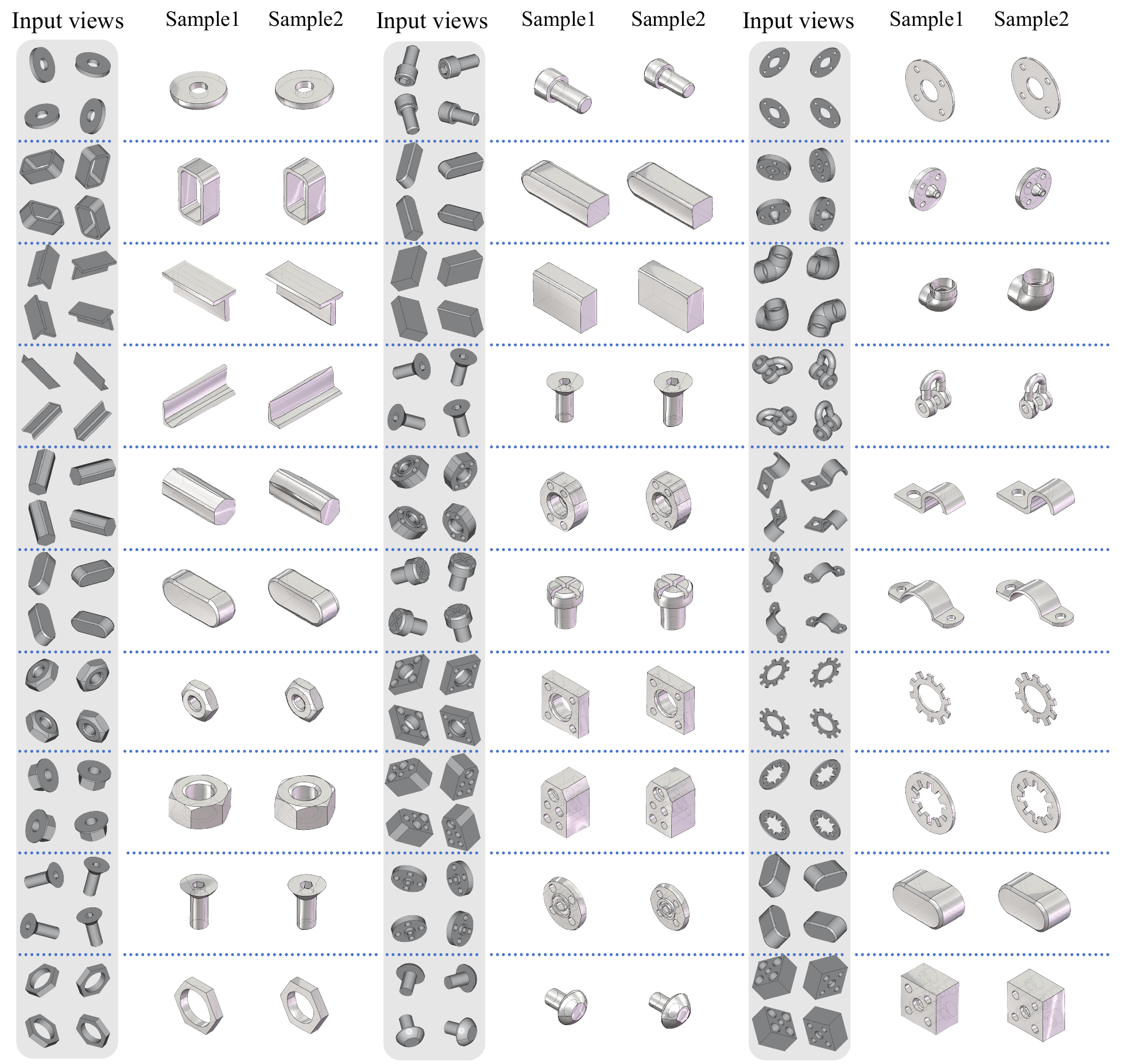}
\caption{\textbf{Multi-view conditioned generation.}
With several views, the outputs align across viewpoints and preserve more fine details.}
\label{fig:cond_multiview}
\end{figure*}

\paragraph{Multi-view conditioned generation}
Fig.~\ref{fig:cond_multiview} shows the results from multiple views. The generations are consistent across viewpoints and recover fine details more reliably than single view inputs.

\section{Unconditional Generations on ABC}
We provide additional unconditional results on the ABC dataset to complement the main experiments. The examples illustrate that our single–stage decoding produces coherent solids with consistent edge–face incidence and fewer topology errors, aligning with the validity gains reported in Sec.~\ref{sec:exp:main}.

\begin{figure*}[t]
\centering
\includegraphics[width=\textwidth]{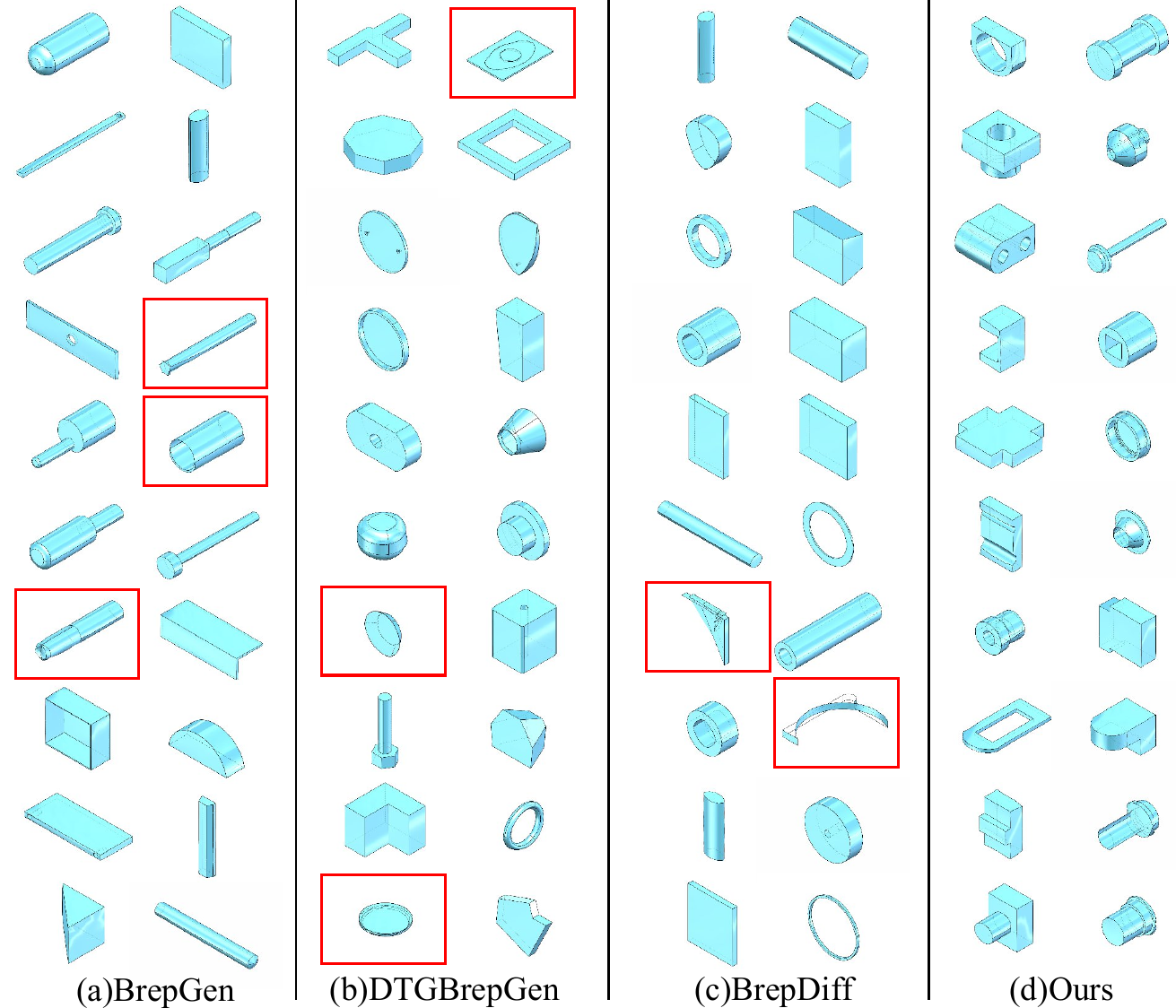}
\caption{\textbf{Unconditional generations on ABC.}
Columns (a–d) show BRepGen, DTGBrepGen, BrepDiff, and \modelName\, respectively. Red boxes highlight typical artifacts observed in baselines, including open shells or missing faces around holes and fillets, non{-}manifold junctions, and degenerate sliver plates or rods.}
\label{fig:abc_uncond}
\end{figure*}


\section{Data Preprocessing and Postprocessing}
We follow the general pipeline of BRepGen~\cite{xu2024brepgen} and DTGBrepGen~\cite{li2025dtgbrepgen} and adapt it to our representation. We first remove periodic seams by splitting all torus faces and ring edges 
so that no periodic faces or edges remain. We then normalize each model to a cube \([-1,1]^3\) centered at the origin. For each face we record its position box \(F_p\) as the two diagonal corners of the face bounding box. For each edge we record its position box \(E_p\) in the same way and we take its two endpoints as \(\mathcal{V}\). Next we normalize every face and every edge again to \([-1,1]^3\) in its own local frame and treat them as B\text{-}spline primitives. A face is represented by a \(6\times6\) control grid \(F_z\). An edge is represented by six control points \(E_z\). In \(F_p\), \(E_p\), and \(\mathcal{V}\), the first point is the minimum corner and the second point is the maximum corner. Each shape becomes a sequence of faces and a sequence of edges. We sort the face sequence by \(F_p\) in lexicographic order and sort the edge sequence by \(E_p\) in the same way. Unlike BRepGen, which stores sampled points, we store control points. This reduces the parameter count and improves surface smoothness, which is consistent with the benefit of control point representations reported by DTGBrepGen.

In postprocessing we reconstruct loops and assemble a solid from the predicted geometry and topology. We start from the predicted edge–face incidence and collect for each face the set of incident edges. Inside a face we try to build one or more closed non intersecting loops by pairing each edge endpoint with the nearest endpoint that belongs to a different edge. The preprocessing step guarantees that there is no ring edge, so a loop corner must connect two different edges. If we obtain valid loops we accept them as the vertex–edge topology for that face. If loop building fails the current sample is marked invalid. We keep two vertex sets during this step. One set is \(\mathcal{V}\). The other set is obtained by de{-}normalizing the two end samples of each edge. We try both and take the one that succeeds. This fallback is the same as in BRepGen.

After we obtain edge–face and vertex–edge topology we compute uniform samples from the predicted control points. We sample each face on a \(32\times32\) grid from \(F_z\) and each edge with 32 points from \(E_z\). We then de{-}normalize faces and edges using \(F_p\) and \(E_p\). We refine the geometry with the joint fitting procedure of BRepGen so that topologically connected surfaces and curves meet cleanly. With the refined curves, surfaces, and all topological relations, we use the OpenCascade kernel to build the final B\text{-}rep. Any exceptions during this step are treated as a compile failure. A sample is considered valid only if the kernel can construct the model, confirm that it is closed, and compute a finite volume. If any of these checks fails the sample is invalid. These checks certify closure and watertightness. Topological legality is already enforced before a model can compile: every edge is incident to exactly two faces and two vertices, each face loop closes head to tail, and no extraneous vertices or edges are present. The extra step of sampling from control points and fitting curves and surfaces explains why our postprocessing is slightly slower than BRepGen even though both pipelines share the same assembly procedure.

\begin{table}[!]
\centering
\caption{\textbf{Maximum shared edges per face pair in ABC (counts).} Left: before splitting periodic faces/edges. Right: after splitting. The distribution is long tailed, so multiple shared edges are common.}
\label{tab:abc_max_shared}
\begin{minipage}{0.48\linewidth}
\centering
\begin{tabular}{cc}
\toprule
max shared & count \\
\midrule
0  & 847   \\
1  & 142020\\
2  & 40735 \\
3  & 1830  \\
4  & 509   \\
5  & 110   \\
6  & 47    \\
7  & 15    \\
8  & 22    \\
9  & 4     \\
10 & 5     \\
11 & 2     \\
12 & 2     \\
\bottomrule
\end{tabular}\\
\smallskip
\textit{Before splitting}
\end{minipage}
\hfill
\begin{minipage}{0.48\linewidth}
\centering
\begin{tabular}{cc}
\toprule
max shared & count \\
\midrule
1  & 68188  \\
2  & 114404 \\
3  & 2736   \\
4  & 737    \\
5  & 48     \\
6  & 26     \\
7  & 2      \\
8  & 4      \\
9  & 1      \\
11 & 1      \\
16 & 1      \\
\bottomrule
\end{tabular}\\
\smallskip
\textit{After splitting}
\end{minipage}
\end{table}

\section{Generality of the Proposed Generator}
We discuss the ability to handle common B\text{-}rep cases where two faces share more than one edge, and we compare the modeling assumptions with HoLa~\cite{liu2025hola} and DTGBrepGen~\cite{li2025dtgbrepgen}. HoLa is an excellent nearly one{-}stage method. It first performs an implicit intersection test between face pairs and then generates the corresponding half{-}edge pairs. This design ties edge creation to a single intersection per face pair. In practice it means one shared edge per face pair unless extra logic is added. Our approach predicts a set of unique edges and a set of unique faces and then assigns each edge to its two incident faces with an explicit incidence matrix. There is no built{-}in cap on how many edges two faces may share. DTGBrepGen adopts a fixed cap of five shared edges per face pair, which in our statistics covers 99.9\% of ABC. The cap is practical, yet the distribution is long tailed, so a cap can become a modeling limit in rare but valid cases.

We measure the maximum number of shared edges between any face pair in each ABC model. The phenomenon is common rather than rare. After splitting periodic faces and ring edges at seams, about 63.37\% of models have a maximum shared count greater than one. Without the split, the share is still about 23.71\%. Tab.~\ref{tab:abc_max_shared} lists the raw counts to show the tail.
In short, modeling edges and faces as separate sets with explicit incidence lets the generator cover these frequent multi{-}edge cases without special rules, while keeping the decoding stage single step and validity aware.

\end{document}